%% file: surveyML_arXiv.tex
\documentclass[twoside,11pt]{article}

\usepackage{jmlr2e_arXiv}
\usepackage{graphicx}
\usepackage{amssymb}
\usepackage{color}
\usepackage{psfrag}
\usepackage{amsmath}
\usepackage{url}
\usepackage{rotating}
\usepackage{multirow}
\usepackage{pifont}
\usepackage{ulem}
\usepackage{soul}
\usepackage{pdflscape}
\usepackage{afterpage}
\graphicspath{ {images/} }

\newcommand{\fref}[1]{Figure~\ref{#1}}
\newcommand{\tref}[1]{Table~\ref{#1}}

\newcommand{\cref}[1]{Chapter~\ref{#1}}
\newcommand{\sref}[1]{Section~\ref{#1}}

\usepackage{cleveref}
\crefformat{footnote}{#2\footnotemark[#1]#3}

\newcommand{\mystar}{\ding{73}}
\newcommand{\mystarfill}{\ding{72}}

\def\argmin{\operatornamewithlimits{arg\,min}}

\newcommand{\innerp}[1]{\left\langle #1 \right\rangle}
\DeclareMathOperator{\rank}{rank}
\DeclareMathOperator{\trace}{tr}
\DeclareMathOperator{\vectorize}{vec}

\usepackage{booktabs} % better tables

\DeclareMathOperator{\sign}{sign}

\jmlrheading{}{}{}{}{}{Aur\'elien Bellet, Amaury Habrard and Marc Sebban}

\ShortHeadings{A Survey on Metric Learning for Feature Vectors and Structured Data}{Bellet, Habrard and Sebban}
\firstpageno{1}

\begin{document}

\title{A Survey on Metric Learning for Feature Vectors and Structured Data}

\author{\name{Aur\'elien Bellet}\thanks{Most of the work in this paper was carried out while the author was affiliated with Laboratoire Hubert Curien UMR 5516, Universit\'e de Saint-Etienne, France.} \email bellet@usc.edu \\
       \addr Department of Computer Science\\
       University of Southern California\\
       Los Angeles, CA 90089, USA\\
       \AND
       \name Amaury Habrard \email amaury.habrard@univ-st-etienne.fr \\
       \name Marc Sebban \email marc.sebban@univ-st-etienne.fr \\
       \addr Laboratoire Hubert Curien UMR 5516\\
       Universit\'e de Saint-Etienne\\
       18 rue Benoit Lauras, 42000 St-Etienne, France}

\editor{}

\authorpdf{A. Bellet, A. Habrard and M. Sebban}

\maketitle

\begin{abstract}%   <- trailing '%' for backward compatibility of .sty file
The need for appropriate ways to measure the distance or similarity between data is ubiquitous in machine learning, pattern recognition and data mining, but handcrafting such good metrics for specific problems is generally difficult. This has led to the emergence of metric learning, which aims at automatically learning a metric from data and has attracted a lot of interest in machine learning and related fields for the past ten years. This survey paper proposes a systematic review of the metric learning literature, highlighting the pros and cons of each approach. We pay particular attention to Mahalanobis distance metric learning, a well-studied and successful framework, but additionally present a wide range of methods that have recently emerged as powerful alternatives, including nonlinear metric learning, similarity learning and local metric learning. Recent trends and extensions, such as semi-supervised metric learning, metric learning for histogram data and the derivation of generalization guarantees, are also covered. Finally, this survey addresses metric learning for structured data, in particular edit distance learning, and attempts to give an overview of the remaining challenges in metric learning for the years to come.
\end{abstract}

\begin{keywords}
  Metric Learning, Similarity Learning, Mahalanobis Distance, Edit Distance 
\end{keywords}

\input{intro}

\input{properties}

\input{maha}

\input{other_numerical}

\input{structured}

\input{conclu}

% Acknowledgements should go at the end, before appendices and references

\acks{We would like to acknowledge support from the ANR LAMPADA 09-EMER-007-02 project.}

\normalem
\bibliography{surveyML}

\end{document}

%% file: intro.tex
\section{Introduction}

%[REPRENDRE DES IDEES DES SLIDES SURVEY MARC]

The notion of {\it pairwise metric}---used throughout this survey as a generic term for distance, similarity or dissimilarity function---between data points plays an important role in many machine learning, pattern recognition and data mining techniques.\footnote{Metric-based learning methods were the focus of the recent SIMBAD European project (ICT 2008-FET 2008-2011). Website: \url{http://simbad-fp7.eu/}} For instance, in classification, the $k$-Nearest Neighbor classifier \citep{Cover1967} uses a metric to identify the nearest neighbors; many clustering algorithms, such as the prominent $K$-Means \citep{Lloyd1982}, rely on distance measurements between data points; in information retrieval, documents are often ranked according to their relevance to a given query based on similarity scores. Clearly, the performance of these methods depends on the quality of the metric: as in the saying ``birds of a feather flock together'', we hope that it identifies as similar (resp. dissimilar) the pairs of instances that are indeed semantically close (resp. different).
General-purpose metrics exist (e.g., the Euclidean distance and the cosine similarity for feature vectors or the Levenshtein distance for strings) but they often fail to capture the idiosyncrasies of the data of interest. Improved results are expected when the metric is designed specifically for the task at hand. Since manual tuning is difficult and tedious, a lot of effort has gone into {\it metric learning}, the research topic devoted to automatically learning metrics from data.

\subsection{Metric Learning in a Nutshell}

Although its origins can be traced back to some earlier work \citep[e.g.,][]{Short1981,Fukunaga1990,Friedman1994,Hastie1996,Baxter1997}, metric learning really emerged in 2002 with the pioneering work of \citet{Xing2002} that formulates it as a convex optimization problem. It has since been a hot research topic, being the subject of tutorials at ICML 2010\footnote{\url{http://www.icml2010.org/tutorials.html}} and ECCV 2010\footnote{\url{http://www.ics.forth.gr/eccv2010/tutorials.php}} and workshops at ICCV 2011,\footnote{\url{http://www.iccv2011.org/authors/workshops/}} NIPS 2011\footnote{\url{http://nips.cc/Conferences/2011/Program/schedule.php?Session=Workshops}} and ICML~2013.\footnote{\url{http://icml.cc/2013/?page_id=41}}

\begin{figure}[t]
\centering
\includegraphics[width=0.95\textwidth]{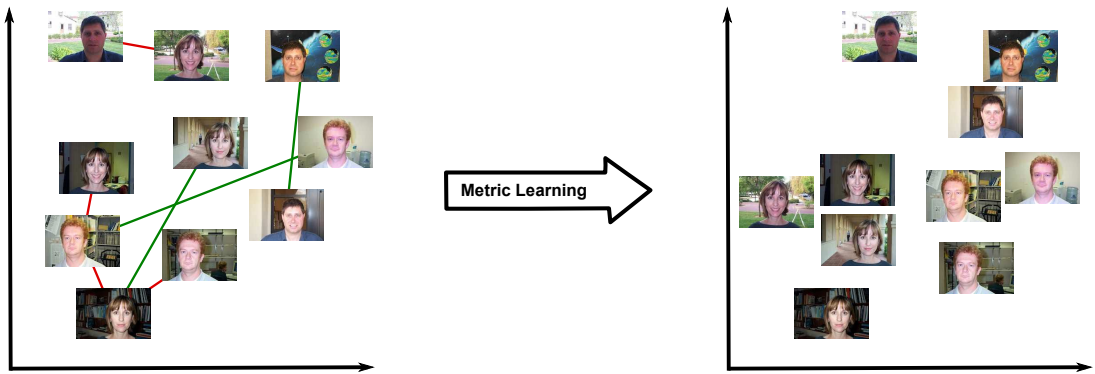}
\caption[Caption for LOF]{Illustration of metric learning applied to a face recognition task. For simplicity, images are represented as points in 2 dimensions. Pairwise constraints, shown in the left pane, are composed of images representing the same person (must-link, shown in green) or different persons (cannot-link, shown in red). We wish to adapt the metric so that there are fewer constraint violations (right pane). Images are taken from the Caltech Faces dataset.\footnotemark }
\label{fig:ml}
\end{figure}

The goal of metric learning is to adapt some pairwise real-valued metric function, say the Mahalanobis distance $d_{\boldsymbol{M}}(\boldsymbol{x},\boldsymbol{x'}) = \sqrt{(\boldsymbol{x}-\boldsymbol{x'})^T\boldsymbol{M}(\boldsymbol{x}-\boldsymbol{x'})}$, to the problem of interest using the information brought by training examples. Most methods learn the metric (here, the positive semi-definite matrix $\boldsymbol{M}$ in $d_{\boldsymbol{M}}$) in a weakly-supervised way from pair or triplet-based constraints of the following form:
\begin{itemize}
\item Must-link / cannot-link constraints (sometimes called positive / negative pairs):
\begin{eqnarray*}
\mathcal{S} & = & \{(x_i,x_j) : x_i\text{ and }x_j\text{ should be similar}\},\\
\mathcal{D} & = & \{(x_i,x_j) : x_i\text{ and }x_j\text{ should be dissimilar}\}.
\end{eqnarray*}
\item Relative constraints (sometimes called training triplets):
$$\mathcal{R} = \{(x_i,x_j,x_k) : x_i\text{ should be more similar to }x_j\text{ than to }x_k\}.$$
\end{itemize}
%\red{In the} semi-supervised \red{setting}, metric learning algorithms are able to make use of additional data points for which no such constraints are available \red{(en dire un peu plus sur leur exploitation)}.
A metric learning algorithm basically aims at finding the parameters of the metric such that it best agrees with these constraints (see \fref{fig:ml} for an illustration), in an effort to approximate the underlying semantic metric. This is typically formulated as an optimization problem that has the following general form:
\begin{eqnarray*}
\begin{aligned}
\min_{\boldsymbol{M}} &&& \ell(\boldsymbol{M},\mathcal{S},\mathcal{D},\mathcal{R})\quad+\quad \lambda R(\boldsymbol{M})%\\
%\text{s.t.} &&& \displaystyle\sum_{(\boldsymbol{x_i},\boldsymbol{x_j})\in \mathcal{S}}d_\boldsymbol{M}^2(\boldsymbol{x_i},\boldsymbol{x_j}) \leq 1.
\end{aligned}
\end{eqnarray*}
where $\ell(\boldsymbol{M},\mathcal{S},\mathcal{D},\mathcal{R})$ is a loss function that incurs a penalty when training constraints are violated, $R(\boldsymbol{M})$ is some regularizer on the parameters $\boldsymbol{M}$ of the learned metric and $\lambda\geq 0$ is the regularization parameter. As we will see in this survey, state-of-the-art metric learning formulations essentially differ by their choice of metric, constraints, loss function and regularizer.

\footnotetext{\url{http://www.vision.caltech.edu/html-files/archive.html}}

\begin{figure}[t]
\centering
\includegraphics[width=0.95\textwidth]{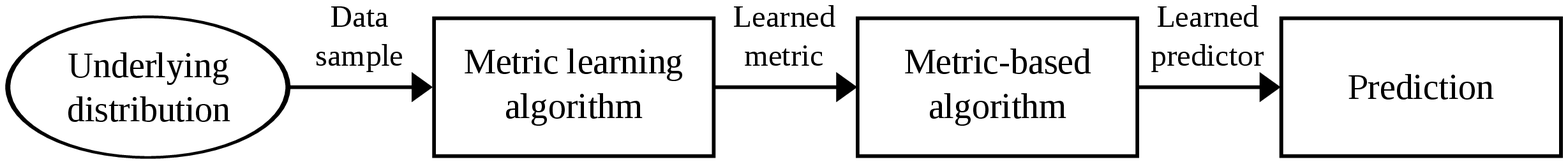}
\caption{The common process in metric learning. A metric is learned from training data and plugged into an algorithm that outputs a predictor (e.g., a classifier, a regressor, a recommender system...) which hopefully performs better than a predictor induced by a standard (non-learned) metric.}
\label{fig:metriclearning}
\end{figure}

After the metric learning phase, the resulting function is used to improve the performance of a metric-based algorithm, which is most often $k$-Nearest Neighbors ($k$-NN), but may also be a clustering algorithm such as $K$-Means, a ranking algorithm, etc. The common process in metric learning is summarized in \fref{fig:metriclearning}.

\subsection{Applications}

Metric learning can potentially be beneficial whenever the notion of metric between instances plays an important role. Recently, it has been applied to problems as diverse as link prediction in networks \citep{Shaw2011}, state representation in reinforcement learning \citep{Taylor2011}, music recommendation \citep{McFee2012}, partitioning problems \citep{Lajugie2014}, identity verification \citep{Ben2012}, webpage archiving \citep{Law2012}, cartoon synthesis \citep{Yu2012} and even assessing the efficacy of acupuncture \citep{Liang2012}, to name a few.
In the following, we list three large fields of application where metric learning has been shown to be very useful.

\paragraph{Computer vision} There is a great need of appropriate metrics in computer vision, not only to compare images or videos in ad-hoc representations---such as bags-of-visual-words \citep{Li2005}---but also in the pre-processing step consisting in building this very representation (for instance, visual words are usually obtained by means of clustering). For this reason, there exists a large body of metric learning literature dealing specifically with computer vision problems, such as image classification \citep{Mensink2012}, object recognition \citep{Frome2007,Verma2012}, face recognition \citep{Guillaumin2009,Lu2012}, visual tracking \citep{Li2012,Jiang2012} or image annotation \citep{Guillaumin2009a}.

\paragraph{Information retrieval} The objective of many information retrieval systems, such as search engines, is to provide the user with the most relevant documents according to his/her query. This ranking is often achieved by using a metric between two documents or between a document and a query. Applications of metric learning to these settings include the work of \citet{Lebanon2006,Lee2008,McFee2010,Lim2013}.

\paragraph{Bioinformatics} Many problems in bioinformatics involve comparing sequences such as DNA, protein or temporal series. These comparisons are based on structured metrics such as edit distance measures (or related string alignment scores) for strings or Dynamic Time Warping distance for temporal series. Learning these metrics to adapt them to the task of interest can greatly improve the results. Examples include the work of \citet{Xiong2006,Saigo2006,Kato2010,Wang2012a}.

\subsection{Related Topics}

We mention here three research topics that are related to metric learning but outside the scope of this survey.

\paragraph{Kernel learning} While metric learning is parametric (one learns the parameters of a given form of metric, such as a Mahalanobis distance), kernel learning is usually nonparametric: one learns the kernel matrix without any assumption on the form of the kernel that implicitly generated it. These approaches are thus very powerful but limited to the transductive setting and can hardly be applied to new data. The interested reader may refer to the recent survey on kernel learning by \citet{Abbasnejad2012}.

\paragraph{Multiple kernel learning} Unlike kernel learning, Multiple Kernel Learning (MKL) is parametric: it learns a combination of predefined base kernels. In this regard, it can be seen as more restrictive than metric or kernel learning, but as opposed to kernel learning, MKL has very efficient formulations and can be applied in the inductive setting. The interested reader may refer to the recent survey on MKL by \citet{Gonen2011}.

\paragraph{Dimensionality reduction} Supervised dimensionality reduction aims at finding a low-dimensional representation that maximizes the separation of labeled data and in this respect has connections with metric learning,\footnote{Some metric learning methods can be seen as finding a new feature space, and a few of them actually have the additional goal of making this feature space low-dimensional.} although the primary objective is quite different. Unsupervised dimensionality reduction, or manifold learning, usually assume that the (unlabeled) data lie on an embedded low-dimensional manifold within the higher-dimensional space and aim at ``unfolding'' it. These methods aim at capturing or preserving some properties of the original data (such as the variance or local distance measurements) in the low-dimensional representation.\footnote{These approaches are sometimes referred to as ``unsupervised metric learning'', which is somewhat misleading because they do not optimize a notion of metric.} The interested reader may refer to the surveys by \citet{Fodor2002} and \citet{Maaten2009}.

\subsection{Why this Survey?}

As pointed out above, metric learning has been a hot topic of research in machine learning for a few years and has now reached a considerable level of maturity both practically and theoretically. The early review due to \citet{Yang2006} is now largely outdated as it misses out on important recent advances: more than 75\% of the work referenced in the present survey is post 2006. A more recent survey, written independently and in parallel to our work, is due to \citet{Kulis2012}. Despite some overlap, it should be noted that both surveys have their own strengths and complement each other well. Indeed, the survey of Kulis takes a more general approach, attempting to provide a unified view of a few core metric learning methods. It also goes into depth about topics that are only briefly reviewed here, such as kernelization, optimization methods and applications. On the other hand, the present survey is a detailed and comprehensive review of the existing literature, covering more than 50 approaches (including many recent works that are missing from Kulis' paper) with their relative merits and drawbacks. Furthermore, we give particular attention to topics that are not covered by Kulis, such as metric learning for structured data and the derivation of generalization guarantees.

We think that the present survey may foster novel research in metric learning and be useful to a variety of audiences, in particular: (i) machine learners wanting to get introduced to or update their knowledge of metric learning will be able to quickly grasp the pros and cons of each method as well as the current strengths and limitations of the research area as a whole, and (ii) machine learning practitioners interested in applying metric learning to their own problem will find information to help them choose the methods most appropriate to their needs, along with links to source codes whenever available.

Note that we focus on general-purpose methods, i.e., that are applicable to a wide range of application domains. The abundant literature on metric learning designed specifically for computer vision is not addressed because the understanding of these approaches requires a significant amount of background in that area. For this reason, we think that they deserve a separate survey, targeted at the computer vision audience.

%[COMME DANS MKL, DONNER LES CHIFFRES, DU GENRE ON DECRIT X APPROCHES OU ON MENTIONNE X REFERENCES?]

\subsection{Prerequisites}
This survey is almost self-contained and has few prerequisites. For metric learning from feature vectors, we assume that the reader has some basic knowledge of linear algebra and convex optimization \citep[if needed, see][for a brush-up]{Boyd2004}.
For metric learning from structured data, we assume that the reader has some familiarity with basic probability theory, statistics and likelihood maximization.
The notations used throughout this survey are summarized in \tref{tab:notations}.

\begin{table}[t]
\centering
\begin{scriptsize}
\begin{tabular}{ll}
\toprule
\textbf{Notation} & \textbf{Description}\\
\midrule
$\mathbb{R}$ & Set of real numbers\\
$\mathbb{R}^d$ & Set of $d$-dimensional real-valued vectors\\
$\mathbb{R}^{c\times d}$ & Set of $c\times d$ real-valued matrices\\
 $\mathbb{S}^{d}_+$ & Cone of symmetric PSD $d\times d$ real-valued matrices\\
 ${\cal X}$ & Input (instance) space\\
 ${\cal Y}$ & Output (label) space\\
 ${\cal S}$ & Set of must-link constraints\\
 ${\cal D}$ & Set of cannot-link constraints\\
 ${\cal R}$ & Set of relative constraints\\
 $z=(x,y)\in\mathcal{X}\times\mathcal{Y}$ & An arbitrary labeled instance\\
 $\boldsymbol{x}$ & An arbitrary vector\\
 $\boldsymbol{M}$ & An arbitrary matrix\\
 $\boldsymbol{I}$ & Identity matrix\\
  $\boldsymbol{M} \succeq 0$ & PSD matrix $\boldsymbol{M}$\\
 $\|\cdot\|_p$ & $p$-norm\\
$\|\cdot\|_\mathcal{F}$ & Frobenius norm\\
$\|\cdot\|_*$ & Nuclear norm\\
$\trace(\boldsymbol{M})$ & Trace of matrix $\boldsymbol{M}$\\
$[t]_ +=\max(0,1-t)$ & Hinge loss function\\
$\xi$ & Slack variable\\
 $\Sigma$ & Finite alphabet\\
$\mathsf{x}$& String of finite size\\
\bottomrule
\end{tabular}
\end{scriptsize}
\caption{Summary of the main notations.}
\label{tab:notations}
\end{table}

\subsection{Outline}

The rest of this paper is organized as follows. We first assume that data consist of vectors lying in some feature space $\mathcal{X}\subseteq\mathbb{R}^d$. \sref{sec:properties} describes key properties that we will use to provide a taxonomy of metric learning algorithms. In \sref{sec:maha}, we review the large body of work dealing with supervised Mahalanobis distance learning. \sref{sec:recentml} deals with recent advances and trends in the field, such as linear similarity learning, nonlinear and local methods, histogram distance learning, the derivation of generalization guarantees and semi-supervised metric learning methods. We cover metric learning for structured data in \sref{sec:structured}, with a focus on edit distance learning. Lastly, we conclude this survey in \sref{sec:conclu} with a discussion on the current limitations of the existing literature and promising directions for future research.

%% file: properties.tex
\section{Key Properties of Metric Learning Algorithms}
\label{sec:properties}

\begin{figure}[t]
\centering
\includegraphics[width=0.95\textwidth]{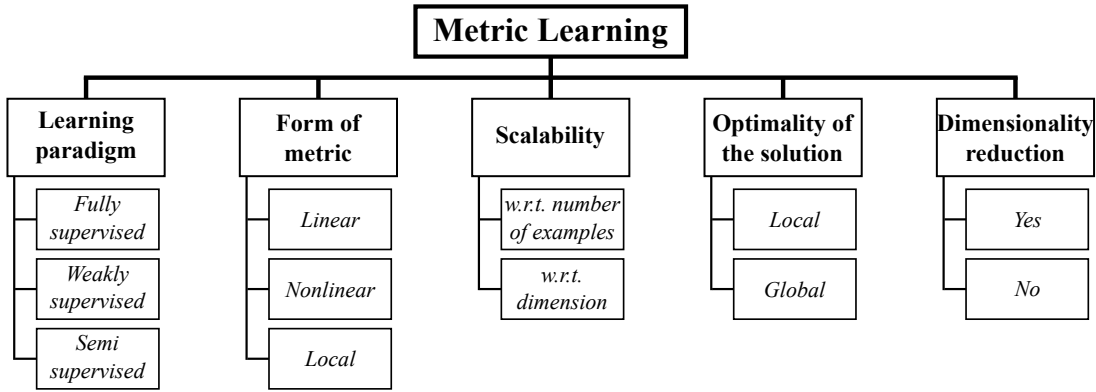}
\caption{Five key properties of metric learning algorithms.}
\label{fig:properties}
\end{figure}

Except for a few early methods, most metric learning algorithms are essentially ``competitive'' in the sense that they are able to achieve state-of-the-art performance on some problems. However, each algorithm has its intrinsic properties (e.g., type of metric, ability to leverage unsupervised data, good scalability with dimensionality, generalization guarantees, etc) and emphasis should be placed on those when deciding which method to apply to a given problem.
In this section, we identify and describe five key properties of metric learning algorithms, summarized in \fref{fig:properties}. We use them to provide a taxonomy of the existing literature: the main features of each method are given in \tref{tab:mlvectsum}.\footnote{Whenever possible, we use the acronyms provided by the authors of the studied methods. When there is no known acronym, we take the liberty of choosing one.}
%More precisely, we will use the learning paradigm (supervised or semi-supervised) as the top-level category, and use the other properties as finer-grained groups whenever appropriate. [VIRER CETTE PHRASE?] More specific properties are also stated when presenting each method.
 
\paragraph{Learning Paradigm}

We will consider three learning paradigms:
\begin{itemize}
\item {\it Fully supervised}: the metric learning algorithm has access to a set of labeled training instances $\{z_i=(x_i,y_i)\}_{i=1}^n$, where each training example $z_i\in\mathcal{Z} = \mathcal{X}\times\mathcal{Y}$ is composed of an instance $x_i\in\mathcal{X}$ and a label (or class) $y_i\in\mathcal{Y}$. $\mathcal{Y}$ is a discrete and finite set of $|\mathcal{Y}|$ labels (unless stated otherwise). In practice, the label information is often used to generate specific sets of pair/triplet constraints $\mathcal{S},\mathcal{D},\mathcal{R}$, for instance based on a notion of neighborhood.\footnote{These constraints are usually derived from the labels prior to learning the metric and never challenged. Note that \citet{Wang2012} propose a more refined (but costly) approach to the problem of building the constraints from labels. Their method alternates between selecting the most relevant constraints given the current metric and learning a new metric based on the current constraints.}
\item {\it Weakly supervised}: the algorithm has no access to the labels of individual training instances: it is only provided with side information in the form of sets of constraints $\mathcal{S},\mathcal{D},\mathcal{R}$. This is a meaningful setting in a variety of applications where labeled data is costly to obtain while such side information is cheap: examples include users' implicit feedback (e.g., clicks on search engine results), citations among articles or links in a network. This can be seen as having label information only at the pair/triplet level.
\item {\it Semi-supervised}: besides the (full or weak) supervision, the algorithm has access to a (typically large) sample of unlabeled instances for which no side information is available. This is useful to avoid overfitting when the labeled data or side information are scarce.
\end{itemize}

\paragraph{Form of Metric}

Clearly, the form of the learned metric is a key choice. One may identify three main families of metrics:
\begin{itemize}
\item {\it Linear metrics}, such as the Mahalanobis distance. Their expressive power is limited but they are easier to optimize (they usually lead to convex formulations, and thus global optimality of the solution) and less prone to overfitting.
\item {\it Nonlinear metrics}, such as the $\chi^2$ histogram distance. They often give rise to nonconvex formulations (subject to local optimality) and may overfit, but they can capture nonlinear variations in the data.
\item {\it Local metrics}, where multiple (linear or nonlinear) local metrics are learned (typically simultaneously) to better deal with complex problems, such as heterogeneous data. They are however more prone to overfitting than global methods since the number of parameters they learn can be very large.
\end{itemize}

\paragraph{Scalability}

With the amount of available data growing fast, the problem of scalability arises in all areas of machine learning. First, it is desirable for a metric learning algorithm to scale well with the number of training examples $n$ (or constraints). As we will see, learning the metric in an online way is one of the solutions. Second, metric learning methods should also scale reasonably well with the dimensionality $d$ of the data. However, since metric learning is often phrased as learning a $d\times d$ matrix, designing algorithms that scale reasonably well with this quantity is a considerable challenge.

\paragraph{Optimality of the Solution}

This property refers to the ability of the algorithm to find the parameters of the metric that satisfy best the criterion of interest. Ideally, the solution is guaranteed to be the {\it global optimum}---this is essentially the case for convex formulations of metric learning. On the contrary, for nonconvex formulations, the solution may only be a {\it local optimum}.

\paragraph{Dimensionality Reduction}

As noted earlier, metric learning is sometimes formulated as finding a projection of the data into a new feature space. An interesting byproduct in this case is to look for a low-dimensional projected space, allowing faster computations as well as more compact representations. This is typically achieved by forcing or regularizing the learned metric matrix to be low-rank.

\afterpage{
\begin{landscape}
\begin{table}
\vspace*{-1.6cm}
\centering
\begin{scriptsize}
\begin{tabular}{cccccccccccccc}
\toprule
\multirow{2}{*}{\textbf{Page}} & \multirow{2}{*}{\textbf{Name}} & \multirow{2}{*}{\textbf{Year}} & \textbf{Source} & \multirow{2}{*}{\textbf{Supervision}} & \textbf{Form of} & \multicolumn{2}{c}{\textbf{Scalability}} & \multirow{2}{*}{\textbf{Optimum}} & \textbf{Dimension} & \multirow{2}{*}{\textbf{Regularizer}} & \textbf{Additional}\\
&&& \textbf{Code} && \textbf{Metric} & {\bf w.r.t. $n$} & {\bf w.r.t. $d$} & & \textbf{Reduction} & & \textbf{Information}\\
\midrule
\pageref{par:mmc} & MMC & 2002 & Yes & Weak & Linear & \mystarfill\mystar\mystar & \mystar\mystar\mystar & Global & No  & None & ---\\
\pageref{par:schultz} & S\&J & 2003 & No & Weak & Linear & \mystarfill\mystarfill\mystar & \mystarfill\mystarfill\mystarfill & Global & No & Frobenius norm & ---\\
\pageref{par:nca} & NCA & 2004 & Yes & Full & Linear &  \mystarfill\mystar\mystar & \mystarfill\mystarfill\mystar  & Local & Yes & None & For $k$-NN\\
\pageref{par:mcml} & MCML & 2005 & Yes & Full & Linear &  \mystarfill\mystar\mystar & \mystar\mystar\mystar & Global & No & None & For $k$-NN\\
\pageref{par:lmnn} & LMNN & 2005 & Yes & Full & Linear &\mystarfill\mystarfill\mystar & \mystarfill\mystar\mystar & Global & No & None &For $k$-NN\\
\pageref{par:rca} & RCA & 2003 & Yes & Weak & Linear &\mystarfill\mystarfill\mystar & \mystarfill\mystarfill\mystar& Global & No & None & ---\\
\pageref{par:itml} & ITML & 2007 & Yes & Weak & Linear &\mystarfill\mystar\mystar & \mystarfill\mystarfill\mystar& Global & No & LogDet & Online version\\
\pageref{par:sdml} & SDML & 2009 & No & Weak & Linear &\mystarfill\mystar\mystar & \mystarfill\mystarfill\mystar& Global & No & LogDet+L$_1$ & $n \ll d$\\
\pageref{par:pola} & POLA & 2004 & No & Weak & Linear &\mystarfill\mystarfill\mystarfill & \mystarfill\mystar\mystar& Global & No & None & Online\\
\pageref{par:lego} & LEGO & 2008 & No & Weak & Linear &\mystarfill\mystarfill\mystarfill & \mystarfill\mystarfill\mystar& Global & No & LogDet & Online\\
\pageref{par:rdml} & RDML & 2009 & No & Weak & Linear &\mystarfill\mystarfill\mystarfill & \mystarfill\mystarfill\mystar& Global & No & Frobenius norm & Online\\
\pageref{par:mdml} & MDML & 2012 & No & Weak & Linear &\mystarfill\mystarfill\mystarfill & \mystarfill\mystar\mystar& Global & Yes & Nuclear norm & Online\\
%\pageref{par:bian} & LL/sHL-ML & 2011 & & & Linear &&& Global & No & None & ---\\
\pageref{par:mtlmnn} & mt-LMNN & 2010 & Yes & Full & Linear &\mystarfill\mystarfill\mystar & \mystar\mystar\mystar & Global & No & Frobenius norm & Multi-task\\
\pageref{par:mlcs} & MLCS & 2011 & No & Weak & Linear & \mystarfill\mystar\mystar & \mystarfill\mystarfill\mystar & Local & Yes & N/A & Multi-task\\
\pageref{par:gpml} & GPML & 2012 & No & Weak & Linear &\mystarfill\mystar\mystar & \mystarfill\mystarfill\mystar& Global & Yes & von Neumann & Multi-task\\
\pageref{par:tml} & TML & 2010 & Yes & Weak & Linear &\mystarfill\mystarfill\mystar & \mystarfill\mystarfill\mystar& Global & No & Frobenius norm & Transfer learning\\
\pageref{par:lpml} & LPML & 2006 & No & Weak & Linear &\mystarfill\mystarfill\mystar & \mystarfill\mystarfill\mystar& Global & Yes & $L_1$ norm & ---\\
\pageref{par:sml} & SML & 2009 & No & Weak & Linear &\mystarfill\mystar\mystar & \mystar\mystar\mystar& Global & Yes & $L_{2,1}$ norm & ---\\
\pageref{par:boost} & BoostMetric & 2009 & Yes & Weak & Linear &\mystarfill\mystar\mystar & \mystarfill\mystarfill\mystar& Global & Yes & None & ---\\
\pageref{par:dmlp} & DML-$p$ & 2012 & No & Weak & Linear & \mystarfill\mystar\mystar & \mystarfill\mystar\mystar & Global & No  & None & ---\\
\pageref{par:rml} & RML & 2010 & No & Weak & Linear &\mystarfill\mystarfill\mystar & \mystar\mystar\mystar& Global & No & Frobenius norm & Noisy constraints\\
\pageref{par:mlr} & MLR & 2010 & Yes & Full & Linear &\mystarfill\mystarfill\mystar & \mystar\mystar\mystar& Global & Yes & Nuclear norm & For ranking\\
\pageref{par:sila} & SiLA & 2008 & No & Full & Linear &\mystarfill\mystarfill\mystar & \mystarfill\mystarfill\mystar& N/A & No & None & Online\\
\pageref{par:gcosla} & gCosLA & 2009 & No & Weak & Linear &\mystarfill\mystarfill\mystarfill & \mystar\mystar\mystar& Global & No & None & Online\\
\pageref{par:oasis} & OASIS & 2009 & Yes & Weak & Linear &\mystarfill\mystarfill\mystarfill & \mystarfill\mystarfill\mystar& Global & No & Frobenius norm & Online\\
\pageref{par:sllc} & SLLC & 2012 & No & Full & Linear &\mystarfill\mystarfill\mystar & \mystarfill\mystarfill\mystar& Global & No & Frobenius norm & For linear classif.\\
\pageref{par:rsl} & RSL & 2013 & No & Full & Linear &\mystarfill\mystar\mystar & \mystarfill\mystarfill\mystar& Local & No & Frobenius norm & Rectangular matrix\\
\pageref{par:lsmd} & LSMD & 2005 & No & Weak & Nonlinear &\mystarfill\mystar\mystar & \mystarfill\mystarfill\mystar& Local & Yes & None & ---\\
\pageref{par:nnca} & NNCA & 2007 & No & Full & Nonlinear & \mystarfill\mystar\mystar & \mystarfill\mystarfill\mystar & Local & Yes & Recons. error & ---\\
\pageref{par:svml} & SVML & 2012 & No & Full & Nonlinear &\mystarfill\mystar\mystar & \mystarfill\mystarfill\mystar& Local & Yes & Frobenius norm & For SVM\\
\pageref{par:gblmnn} & GB-LMNN & 2012 & No & Full & Nonlinear &\mystarfill\mystarfill\mystar & \mystarfill\mystarfill\mystar& Local & Yes & None & ---\\
\pageref{par:hdml} & HDML & 2012 & Yes & Weak & Nonlinear & \mystarfill\mystarfill\mystar & \mystarfill\mystarfill\mystar & Local & Yes & $L_2$ norm & Hamming distance\\
\pageref{par:mmlmnn} & M$^2$-LMNN & 2008 & Yes & Full & Local &\mystarfill\mystarfill\mystar & \mystarfill\mystar\mystar& Global & No & None & ---\\
\pageref{par:glml} & GLML & 2010 & No & Full & Local & \mystarfill\mystarfill\mystarfill & \mystarfill\mystarfill\mystar & Global & No & Diagonal & Generative\\
\pageref{par:bkmeans} & Bk-means & 2009 & No & Weak & Local &\mystarfill\mystar\mystar & \mystarfill\mystarfill\mystarfill& Global & No & RKHS norm & Bregman dist.\\
\pageref{par:plml} & PLML & 2012 & Yes & Weak & Local &\mystarfill\mystarfill\mystar & \mystar\mystar\mystar& Global & No & Manifold+Frob & ---\\
\pageref{par:rfd} & RFD & 2012 & Yes & Weak & Local &\mystarfill\mystarfill\mystar & \mystarfill\mystarfill\mystarfill& N/A & No & None & Random forests\\
\pageref{par:chilmnn} & $\chi^2$-LMNN & 2012 & No & Full & Nonlinear &\mystarfill\mystarfill\mystar & \mystarfill\mystarfill\mystar& Local & Yes & None & Histogram data\\
\pageref{par:gml} & GML & 2011 & No & Weak & Linear &\mystarfill\mystar\mystar & \mystarfill\mystarfill\mystar& Local & No & None & Histogram data\\
\pageref{par:emdl} & EMDL & 2012 & No & Weak & Linear &\mystarfill\mystar\mystar & \mystarfill\mystarfill\mystar& Local & No & Frobenius norm & Histogram data\\
\pageref{par:lrml} & LRML & 2008 & Yes & Semi & Linear &\mystarfill\mystar\mystar & \mystar\mystar\mystar& Global & No & Laplacian & ---\\
\pageref{par:lmdml} & M-DML & 2009 & No & Semi & Linear &\mystarfill\mystar\mystar & \mystar\mystar\mystar& Local & No & Laplacian & Auxiliary metrics\\
\pageref{par:seraph} & SERAPH & 2012 & Yes & Semi & Linear & \mystarfill\mystar\mystar & \mystar\mystar\mystar & Local & Yes & Trace+entropy & Probabilistic\\
\pageref{par:cdml} & CDML & 2011 & No & Semi & N/A &N/A & N/A & N/A & N/A & N/A & Domain adaptation\\
\pageref{par:daml} & DAML & 2011 & No & Semi & Nonlinear & \mystarfill\mystar\mystar & \mystar\mystar\mystar & Global & No & MMD & Domain adaptation\\
\bottomrule
\end{tabular}
\end{scriptsize}
\caption{Main features of metric learning methods for feature vectors. Scalability levels are relative and given as a rough guide.}
\label{tab:mlvectsum}
\end{table}
\end{landscape}
}

%% file: maha.tex
\section{Supervised Mahalanobis Distance Learning}
\label{sec:maha}

This section deals with (fully or weakly) supervised Malahanobis distance learning (sometimes simply referred to as distance metric learning), which has attracted a lot of interest due to its simplicity and nice interpretation in terms of a linear projection. We start by presenting the Mahalanobis distance and two important challenges associated with learning this form of metric.

\begin{figure}[t]
\centering
\includegraphics[width=0.5\textwidth]{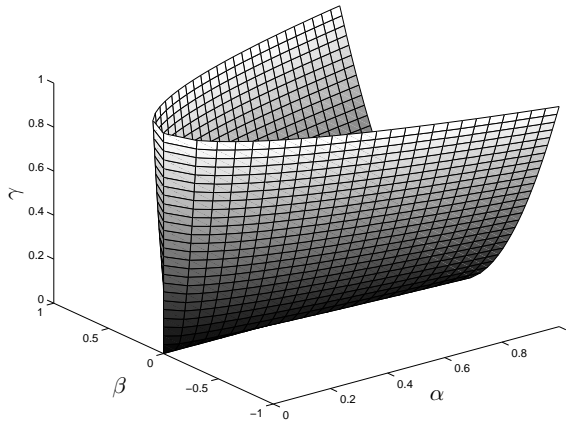}
\caption[The cone $\mathbb{S}^{2}_+$ of positive semi-definite 2x2 matrices.]{The cone $\mathbb{S}^{2}_+$ of positive semi-definite 2x2 matrices of the form
$\begin{bmatrix} \alpha & \beta \\
\beta & \gamma
\end{bmatrix}$.}
\label{fig:psdcone}
\end{figure}

\paragraph{The Mahalanobis distance} This term comes from \citet{Mahalanobis1936} and originally refers to a distance measure that incorporates the correlation between features:
$$d_{maha}(\boldsymbol{x},\boldsymbol{x'}) = \sqrt{(\boldsymbol{x}-\boldsymbol{x'})^T\boldsymbol{\Omega}^{-1}(\boldsymbol{x}-\boldsymbol{x'})},$$
where $\mathbf{x}$ and $\mathbf{x'}$ are random vectors from the same distribution with covariance matrix $\boldsymbol{\Omega}$. By an abuse of terminology common in the metric learning literature, we will in fact use the term Mahalanobis distance to refer to generalized quadratic distances, defined as
$$d_{\boldsymbol{M}}(\boldsymbol{x},\boldsymbol{x'}) = \sqrt{(\boldsymbol{x}-\boldsymbol{x'})^T\boldsymbol{M}(\boldsymbol{x}-\boldsymbol{x'})}$$
and parameterized by $\boldsymbol{M}\in \mathbb{S}^{d}_+$, where $\mathbb{S}^{d}_+$ is the cone of symmetric positive semi-definite (PSD) $d\times d$ real-valued matrices (see \fref{fig:psdcone}).\footnote{Note that in practice, to get rid of the square root, the Mahalanobis distance is learned in its more convenient squared form $d^2_{\boldsymbol{M}}(\boldsymbol{x},\boldsymbol{x'}) = (\boldsymbol{x}-\boldsymbol{x'})^T\boldsymbol{M}(\boldsymbol{x}-\boldsymbol{x'})$.}
$\boldsymbol{M}\in \mathbb{S}^{d}_+$ ensures that $d_{\boldsymbol{M}}$ satisfies the properties of a pseudo-distance: $\forall \boldsymbol{x},\boldsymbol{x'},\boldsymbol{x''}\in\mathcal{X}$,
\begin{enumerate}
\item $d_{\boldsymbol{M}}(\boldsymbol{x},\boldsymbol{x'}) \geq 0\quad$ (nonnegativity),
\item $d_{\boldsymbol{M}}(\boldsymbol{x},\boldsymbol{x}) = 0\quad$ (identity),
\item $d_{\boldsymbol{M}}(\boldsymbol{x},\boldsymbol{x'}) = d(\boldsymbol{x'},\boldsymbol{x})\quad$ (symmetry),
\item $d_{\boldsymbol{M}}(\boldsymbol{x},\boldsymbol{x''}) \leq d(\boldsymbol{x},\boldsymbol{x'})+d(\boldsymbol{x'},\boldsymbol{x''})\quad$ (triangle inequality).
\end{enumerate}

\paragraph{Interpretation} Note that when $\boldsymbol{M}$ is the identity matrix, we recover the Euclidean distance. Otherwise, one can express $\boldsymbol{M}$ as $\boldsymbol{L}^T\boldsymbol{L}$, where $\boldsymbol{L}\in \mathbb{R}^{k\times d}$ where $k$ is the rank of $\boldsymbol{M}$. We can then rewrite $d_{\boldsymbol{M}}(\boldsymbol{x},\boldsymbol{x'})$ as follows:
\begin{eqnarray*}
d_{\boldsymbol{M}}(\boldsymbol{x},\boldsymbol{x'}) & = & \sqrt{(\boldsymbol{x}-\boldsymbol{x'})^T\boldsymbol{M}(\boldsymbol{x}-\boldsymbol{x'})}\\
& = & \sqrt{(\boldsymbol{x}-\boldsymbol{x'})^T\boldsymbol{L}^T\boldsymbol{L}(\boldsymbol{x}-\boldsymbol{x'})}\\
& = & \sqrt{(\boldsymbol{L}\boldsymbol{x}-\boldsymbol{L}\boldsymbol{x'})^T(\boldsymbol{L}\boldsymbol{x}-\boldsymbol{L}\boldsymbol{x'})}.
\end{eqnarray*}
Thus, a Mahalanobis distance implicitly corresponds to computing the Euclidean distance after the linear projection of the data defined by the transformation matrix $\boldsymbol{L}$. Note that if $\boldsymbol{M}$ is low-rank, i.e., $\rank(\boldsymbol{M}) = r < d$, then it induces a linear projection of the data into a space of lower dimension $r$. It thus allows a more compact representation of the data and cheaper distance computations, especially when the original feature space is high-dimensional. These nice properties explain why learning Mahalanobis distance has attracted a lot of interest and is a major component of metric learning.

\paragraph{Challenges} This leads us to two important challenges associated with learning Mahalanobis distances. The first one is to maintain $\boldsymbol{M}\in \mathbb{S}^{d}_+$ in an efficient way during the optimization process. A simple way to do this is to use the projected gradient method which consists in alternating between a gradient step and a projection step onto the PSD cone by setting the negative eigenvalues to zero.\footnote{Note that \citet{Qian2013} have proposed some heuristics to avoid doing this projection at each iteration.} However this is expensive for high-dimensional problems as eigenvalue decomposition scales in $O(d^3)$. The second challenge is to learn a low-rank matrix (which implies a low-dimensional projection space, as noted earlier) instead of a full-rank one. Unfortunately, optimizing $\boldsymbol{M}$ subject to a rank constraint or regularization is NP-hard and thus cannot be carried out efficiently.\\

The rest of this section is a comprehensive review of the supervised Mahalanobis distance learning methods of the literature. We first present two early approaches (\sref{sec:earlyml}). We then discuss methods that are specific to $k$-nearest neighbors (\sref{sec:knnml}), inspired from information theory (\sref{sec:itml}), online learning approaches (\sref{sec:onlineml}), multi-task learning (\sref{sec:mt}) and a few more that do not fit any of the previous categories (\sref{sec:othermaha}).

\subsection{Early Approaches}
\label{sec:earlyml}

The approaches in this section deal with the PSD constraint in a rudimentary way.

\paragraph{MMC (Xing et al.)}
\label{par:mmc}

The seminal work of \citet{Xing2002} is the first Mahalanobis distance learning method.\footnote{Source code available at: \url{http://www.cs.cmu.edu/~epxing/papers/}} It relies on a convex formulation with no regularization, which aims at maximizing the sum of distances between dissimilar points while keeping the sum of distances between similar examples small:
\begin{equation}
\label{eq:xing}
\begin{aligned}
\max_{\boldsymbol{M}\in \mathbb{S}^{d}_+} &&& \displaystyle\sum_{(\boldsymbol{x}_i,\boldsymbol{x}_j)\in \mathcal{D}}d_{\boldsymbol{M}}(\boldsymbol{x}_i,\boldsymbol{x}_j)\\
\text{s.t.} &&& \displaystyle\sum_{(\boldsymbol{x}_i,\boldsymbol{x}_j)\in \mathcal{S}}d_{\boldsymbol{M}}^2(\boldsymbol{x}_i,\boldsymbol{x}_j) \leq 1.
\end{aligned}
\end{equation}
The algorithm used to solve \eqref{eq:xing} is a simple projected gradient approach requiring the full eigenvalue decomposition of $\boldsymbol{M}$ at each iteration. This is typically intractable for medium and high-dimensional problems.

\paragraph{S\&J (Schultz \& Joachims)}
\label{par:schultz}

The method proposed by \citet{Schultz2003} relies on the parameterization $\boldsymbol{M} = \boldsymbol{A}^T\boldsymbol{W}\boldsymbol{A}$, where $\boldsymbol{A}$ is fixed and known and $\boldsymbol{W}$ diagonal. We get:
$$d_{\boldsymbol{M}}^2(\boldsymbol{x}_i,\boldsymbol{x}_j) = (\boldsymbol{A}\boldsymbol{x}_i-\boldsymbol{A}\boldsymbol{x}_j)^T\boldsymbol{W}(\boldsymbol{A}\boldsymbol{x}_i-\boldsymbol{A}\boldsymbol{x}_j).$$
By definition, $\boldsymbol{M}$ is PSD and thus one can optimize over the diagonal matrix $\boldsymbol{W}$ and avoid the need for costly projections on the PSD cone. They propose a formulation based on triplet constraints:
\begin{equation}
\label{eq:schultz}
\begin{aligned}
 \min_{\boldsymbol{W}} &&& \|\boldsymbol{M}\|_{\mathcal{F}}^2\quad+\quad C\sum_{i,j,k}\xi_{ijk}\\
 \text{s.t.} &&& d_{\boldsymbol{M}}^2(\boldsymbol{x}_i,\boldsymbol{x}_k) - d_{\boldsymbol{M}}^2(\boldsymbol{x}_i,\boldsymbol{x}_j) \geq 1 - \xi_{ijk} && \forall (\boldsymbol{x}_i,\boldsymbol{x}_j,\boldsymbol{x}_k)\in \mathcal{R},
\end{aligned}
\end{equation}
where $\|\boldsymbol{M}\|_{\mathcal{F}}^2 = \sum_{i,j} M_{ij}^2$ is the squared Frobenius norm of $\boldsymbol{M}$, the $\xi_{ijk}$'s are ``slack'' variables to allow soft constraints\footnote{This is a classic trick used for instance in soft-margin SVM \citep{Cortes1995}. Throughout this survey, we will consistently use the symbol $\xi$ to denote slack variables.} and $C\geq 0$ is the trade-off parameter between regularization and constraint satisfaction. Problem \eqref{eq:schultz} is convex and can be solved efficiently. The main drawback of this approach is that it is less general than full Mahalanobis distance learning: one only learns a weighting $\boldsymbol{W}$ of the features. Furthermore, $\boldsymbol{A}$ must be chosen manually.

\subsection{Approaches Driven by Nearest Neighbors}
\label{sec:knnml}

The objective functions of the methods presented in this section are related to a nearest neighbor prediction rule.

\paragraph{NCA (Goldberger et al.)}
\label{par:nca}

The idea of Neighbourhood Component Analysis\footnote{Source code available at: \url{http://www.ics.uci.edu/~fowlkes/software/nca/}} (NCA), introduced by \citet{Goldberger2004}, is to optimize the expected leave-one-out error of a stochastic nearest neighbor classifier in the projection space induced by $d_{\boldsymbol{M}}$. They use the decomposition $\boldsymbol{M} = \boldsymbol{L}^T\boldsymbol{L}$ and they define the probability that $\boldsymbol{x}_i$ is the neighbor of $\boldsymbol{x}_j$ by
\begin{eqnarray*}
p_{ij} = \frac{\exp(-\|\boldsymbol{L}\boldsymbol{x}_i-\boldsymbol{L}\boldsymbol{x}_j\|_2^2)}{\sum_{l\neq i}\exp(-\|\boldsymbol{L}\boldsymbol{x}_i-\boldsymbol{L}\boldsymbol{x_l}\|_2^2)}, & p_{ii}=0.
\end{eqnarray*}
Then, the probability that $\boldsymbol{x}_i$ is correctly classified is:
$$p_i = \displaystyle\sum_{j : y_j = y_i}p_{ij}.$$
They learn the distance by solving:
\begin{equation}
\label{eq:nca}
\displaystyle\max_L \sum_i p_i.
\end{equation}
Note that the matrix $\boldsymbol{L}$ can be chosen to be rectangular, inducing a low-rank $\boldsymbol{M}$.
The main limitation of \eqref{eq:nca} is that it is nonconvex and thus subject to local maxima. \citet{Hong2011} later proposed to learn a mixture of NCA metrics, while \citet{Tarlow2013} generalize NCA to $k$-NN with $k>1$.

\paragraph{MCML (Globerson \& Roweis)}
\label{par:mcml}

Shortly after Goldberger et al., \citet{Globerson2005} proposed MCML (Maximally Collapsing Metric Learning), an alternative convex formulation based on minimizing a KL divergence between $p_{ij}$ and an ideal distribution, which can be seen as attempting to collapse each class to a single point.\footnote{An implementation is available within the Matlab Toolbox for Dimensionality Reduction:\\ \url{http://homepage.tudelft.nl/19j49/Matlab_Toolbox_for_Dimensionality_Reduction.html}} Unlike NCA, the optimization is done with respect to the matrix $\boldsymbol{M}$ and the problem is thus convex. However, like MMC, MCML requires costly projections onto the PSD cone.

\paragraph{LMNN (Weinberger et al.)}
\label{par:lmnn}

Large Margin Nearest Neighbors\footnote{Source code available at: \url{http://www.cse.wustl.edu/~kilian/code/code.html}} (LMNN), introduced by Weinberger et al. \citeyearpar{Weinberger2005,Weinberger2008a,Weinberger2009}, is one of the most widely-used Mahalanobis distance learning methods and has been the subject of many extensions (described in later sections). One of the reasons for its popularity is that the constraints are defined in a local way: the $k$ nearest neighbors (the ``target neighbors'') of any training instance should belong to the correct class while keeping away instances of other classes (the ``impostors''). The Euclidean distance is used to determine the target neighbors. Formally, the constraints are defined in the following way:
\begin{eqnarray*}
\mathcal{S} & = & \{(\boldsymbol{x}_i,\boldsymbol{x}_j) : y_i=y_j\text{ and }\boldsymbol{x}_j\text{ belongs to the $k$-neighborhood of }\boldsymbol{x}_i\}, \\
\mathcal{R} & = & \{(\boldsymbol{x}_i,\boldsymbol{x}_j,\boldsymbol{x}_k) : (\boldsymbol{x}_i,\boldsymbol{x}_j)\in \mathcal{S}, y_i\neq y_k\}.
\end{eqnarray*}

The distance is learned using the following convex program:
\begin{equation}
\label{eq:lmnn}
\begin{aligned}
\min_{\boldsymbol{M}\in \mathbb{S}^{d}_+} &&& \displaystyle(1-\mu)\sum_{(\boldsymbol{x}_i,\boldsymbol{x}_j)\in \mathcal{S}}d_{\boldsymbol{M}}^2(\boldsymbol{x}_i,\boldsymbol{x}_j)\quad+\quad \mu\sum_{i,j,k}\xi_{ijk}\\
\text{s.t.} &&& d_{\boldsymbol{M}}^2(\boldsymbol{x}_i,\boldsymbol{x}_k) - d_{\boldsymbol{M}}^2(\boldsymbol{x}_i,\boldsymbol{x}_j) \geq 1 - \xi_{ijk} && \forall (\boldsymbol{x}_i,\boldsymbol{x}_j,\boldsymbol{x}_k)\in \mathcal{R},
\end{aligned}
\end{equation}
where $\mu\in[0,1]$ controls the ``pull/push'' trade-off. The authors developed a special-purpose solver---based on subgradient descent and careful book-keeping---that is able to deal with billions of constraints. Alternative ways of solving the problem have been proposed \citep{Torresani2006,Nguyen2008,Park2011,Der2012}. LMNN generally performs very well in practice, although it is sometimes prone to overfitting due to the absence of regularization, especially in high dimension. It is also very sensitive to the ability of the Euclidean distance to select relevant target neighbors. Note that \citet{Do2012} highlighted a relation between LMNN and Support Vector Machines.

\subsection{Information-Theoretic Approaches}
\label{sec:itml}

The methods presented in this section frame metric learning as an optimization problem involving an information measure.

\paragraph{RCA (Bar-Hillel et al.)}
\label{par:rca}

Relevant Component Analysis\footnote{Source code available at: \url{http://www.scharp.org/thertz/code.html}} \citep{Shental2002,Bar-Hillel2003,Bar-Hillel2005} makes use of positive pairs only and is based on subsets of the training examples called ``chunklets''. These are obtained from the set of positive pairs by applying a transitive closure: for instance, if $(\boldsymbol{x}_1,\boldsymbol{x}_2)\in\mathcal{S}$ and $(\boldsymbol{x}_2,\boldsymbol{x}_3)\in\mathcal{S}$, then $\boldsymbol{x}_1$, $\boldsymbol{x}_2$ and $\boldsymbol{x}_3$ belong to the same chunklet. Points in a chunklet are believed to share the same label. Assuming a total of $n$ points in $k$ chunklets, the algorithm is very efficient since it simply amounts to computing the following matrix:
$$\boldsymbol{\hat{C}} = \frac{1}{n}\sum_{j=1}^{k}\sum_{i=1}^{n_j}(x_{ji}-\hat{m}_j)(x_{ji}-\hat{m}_j)^T,$$
where chunklet $j$ consists of $\{x_{ji}\}_{i=1}^{n_j}$ and $\hat{m}_j$ is its mean. Thus, RCA essentially reduces the within-chunklet variability in an effort to identify features that are irrelevant to the task. The inverse of $\boldsymbol{\hat{C}}$ is used in a Mahalanobis distance. The authors have shown that (i) it is the optimal solution to an information-theoretic criterion involving a mutual information measure, and (ii) it is also the optimal solution to the optimization problem consisting in minimizing the within-class distances. An obvious limitation of RCA is that it cannot make use of the discriminative information brought by negative pairs, which explains why it is not very competitive in practice. RCA was later extended to handle negative pairs, at the cost of a more expensive algorithm \citep{Hoi2006,Yeung2006}.

\paragraph{ITML (Davis et al.)}
\label{par:itml}

Information-Theoretic Metric Learning\footnote{Source code available at: \url{http://www.cs.utexas.edu/~pjain/itml/}} (ITML), proposed by \citet{Davis2007}, is an important work because it introduces LogDet divergence regularization that will later be used in several other Mahalanobis distance learning methods \citep[e.g.,][]{Jain2008,Qi2009}. This Bregman divergence on positive definite matrices is defined as:
$$D_{ld}(\boldsymbol{M},\boldsymbol{M}_0)=\trace(\boldsymbol{M}\boldsymbol{M}_0^{-1})-\log\det(\boldsymbol{M}\boldsymbol{M}_0^{-1})-d,$$
where $d$ is the dimension of the input space and $\boldsymbol{M}_0$ is some positive definite matrix we want to remain close to. In practice, $\boldsymbol{M}_0$ is often set to $\boldsymbol{I}$ (the identity matrix) and thus the regularization aims at keeping the learned distance close to the Euclidean distance. The key feature of the LogDet divergence is that it is finite if and only if $\boldsymbol{M}$ is positive definite. Therefore, minimizing $D_{ld}(\boldsymbol{M},\boldsymbol{M}_0)$ provides an automatic and cheap way of preserving the positive semi-definiteness of $\boldsymbol{M}$. ITML is formulated as follows:
\begin{equation}
\label{eq:itml}
\begin{aligned}
\min_{\boldsymbol{M}\in \mathbb{S}^{d}_+} &&& D_{ld}(\boldsymbol{M},\boldsymbol{M}_0) \quad+\quad \gamma\sum_{i,j}\xi_{ij}\\
\text{s.t.} &&& d_{\boldsymbol{M}}^2(\boldsymbol{x}_i,\boldsymbol{x}_j) \leq u + \xi_{ij} && \forall (\boldsymbol{x}_i,\boldsymbol{x}_j)\in \mathcal{S}\\
 &&& d_{\boldsymbol{M}}^2(\boldsymbol{x}_i,\boldsymbol{x}_j) \geq v - \xi_{ij} && \forall (\boldsymbol{x}_i,\boldsymbol{x}_j)\in \mathcal{D},
\end{aligned}
\end{equation}
where $u,v\in\mathbb{R}$ are threshold parameters and $\gamma\geq 0$ the trade-off parameter. ITML thus aims at satisfying the similarity and dissimilarity constraints while staying as close as possible to the Euclidean distance (if $\boldsymbol{M}_0 = \boldsymbol{I}$). More precisely, the information-theoretic interpretation behind minimizing $D_{ld}(\boldsymbol{M},\boldsymbol{M}_0)$ is that it is equivalent to minimizing the KL divergence between two multivariate Gaussian distributions parameterized by $\boldsymbol{M}$ and $\boldsymbol{M}_0$.
The algorithm proposed to solve \eqref{eq:itml} is efficient, converges to the global minimum and the resulting distance performs well in practice. A limitation of ITML is that $\boldsymbol{M}_0$, that must be picked by hand, can have an important influence on the quality of the learned distance. Note that \citet{Kulis2009a} have shown how hashing can be used together with ITML to achieve fast similarity search.
%Finally, note that $\boldsymbol{M}_0$ can be used to incorporate prior knowledge on the considered task.

\paragraph{SDML (Qi et al.)}
\label{par:sdml}

With Sparse Distance Metric Learning (SDML), \citet{Qi2009} specifically deal with the case of high-dimensional data together with few training samples, i.e., $n \ll d$. To avoid overfitting, they use a double regularization: the LogDet divergence (using $\boldsymbol{M}_0=\boldsymbol{I}$ or $\boldsymbol{M}_0=\boldsymbol{\Omega}^{-1}$ where $\boldsymbol{\Omega}$ is the covariance matrix) and $L_1$-regularization on the off-diagonal elements of $\boldsymbol{M}$.
The justification for using this $L_1$-regularization is two-fold: (i) a practical one is that in high-dimensional spaces, the off-diagonal elements of $\boldsymbol{\Omega}^{-1}$ are often very small, and (ii) a theoretical one suggested by a consistency result from a previous work in covariance matrix estimation \citep{Ravikumar2011} that applies to SDML.
They use a fast algorithm based on block-coordinate descent (the optimization is done over each row of $\boldsymbol{M}^{-1}$) and obtain very good performance for the specific case $n \ll d$.

\subsection{Online Approaches}
\label{sec:onlineml}

In online learning \citep{Littlestone1988}, the algorithm receives training instances one at a time and updates at each step the current hypothesis. Although the performance of online algorithms is typically inferior to batch algorithms, they are very useful to tackle large-scale problems that batch methods fail to address due to time and space complexity issues.
Online learning methods often come with regret bounds, stating that the accumulated loss suffered along the way is not much worse than that of the best hypothesis chosen in hindsight.\footnote{A regret bound has the following general form: $\sum_{t=1}^T \ell(h_t,z_t) - \sum_{t=1}^T \ell(h^*,z_t) \leq O(T)$, where $T$ is the number of steps, $h_t$ is the hypothesis at time $t$ and $h^*$ is the best batch hypothesis.}

\paragraph{POLA (Shalev-Shwartz et al.)}
\label{par:pola}

POLA \citep{Shalev-Shwartz2004}, for Pseudo-metric Online Learning Algorithm, is the first online Mahalanobis distance learning approach and learns the matrix $\boldsymbol{M}$ as well as a threshold $b\geq 1$. At each step $t$, POLA receives a pair $(\boldsymbol{x}_i,\boldsymbol{x}_j,y_{ij})$, where $y_{ij}=1$ if $(\boldsymbol{x}_i,\boldsymbol{x}_j)\in\mathcal{S}$ and $y_{ij}=-1$ if $(\boldsymbol{x}_i,\boldsymbol{x}_j)\in\mathcal{D}$, and performs two successive orthogonal projections:
\begin{enumerate}
\item Projection of the current solution $(\boldsymbol{M}^{t-1},b^{t-1})$ onto the set $C_1 = \{(\boldsymbol{M},b)\in \mathbb{R}^{d^2+1}:[y_{ij}(d_{\boldsymbol{M}}^2(\boldsymbol{x}_i,\boldsymbol{x}_j)-b)+1]_+=0\}$, which is done efficiently (closed-form solution). The constraint basically requires that the distance between two instances of same (resp. different) labels be below (resp. above) the threshold $b$ with a margin 1. We get an intermediate solution $(\boldsymbol{M}^{t-\frac{1}{2}},b^{t-\frac{1}{2}})$ that satisfies this constraint while staying as close as possible to the previous solution.
\item Projection of $(\boldsymbol{M}^{t-\frac{1}{2}},b^{t-\frac{1}{2}})$ onto the set $C_2 = \{(\boldsymbol{M},b)\in \mathbb{R}^{d^2+1}:\boldsymbol{M}\in \mathbb{S}^{d}_+,b\geq1\}$, which is done rather efficiently (in the worst case, one only needs to compute the minimal eigenvalue of $\boldsymbol{M}^{t-\frac{1}{2}}$). This projects the matrix back onto the PSD cone. We thus get a new solution $(\boldsymbol{M}^{t},b^{t})$ that yields a valid Mahalanobis distance.
\end{enumerate}
A regret bound for the algorithm is provided. %However, POLA relies on the unrealistic assumption that there exists $(\boldsymbol{M^*},b^*)$ such that $[y_{ij}(d_{\boldsymbol{M^*}}^2(\boldsymbol{x}_i,\boldsymbol{x}_j)-b^*)+1]_+=0$ for all training pairs (i.e., there exists a matrix and a threshold value that perfectly separate them with margin 1), and is not competitive in practice.

\paragraph{LEGO (Jain et al.)}
\label{par:lego}

LEGO (Logdet Exact Gradient Online), developed by \citet{Jain2008}, is an improved version of POLA based on LogDet divergence regularization. It features tighter regret bounds, more efficient updates and better practical performance.

%\paragraph{ITML (David et al.)}
%\label{par:itml2}

%ITML, presented in \sref{sec:itml}, also has an online version with bounded regret. At each step, the algorithm minimizes a tradeoff between LogDet regularization with respect to the previous matrix and a square loss. The resulting distance generally performs slightly worse than the batch version but the algorithm can be faster.

\paragraph{RDML (Jin et al.)}
\label{par:rdml}

RDML \citep{Jin2009} is similar to POLA in spirit but is more flexible. At each step $t$, instead of forcing the margin constraint to be satisfied, it performs a gradient descent step of the following form (assuming Frobenius regularization):
$$\boldsymbol{M}^{t} = \pi_{\mathbb{S}^{d}_+}\left(\boldsymbol{M}^{t-1} - \lambda y_{ij}(\boldsymbol{x}_i-\boldsymbol{x}_j)(\boldsymbol{x}_i-\boldsymbol{x}_j)^T\right),$$
where $\pi_{\mathbb{S}^{d}_+}(\cdot)$ is the projection to the PSD cone. The parameter $\lambda$ implements a trade-off between satisfying the pairwise constraint and staying close to the previous matrix $\boldsymbol{M}^{t-1}$. Using some linear algebra, the authors show that this update can be performed by solving a convex quadratic program instead of resorting to eigenvalue computation like POLA. RDML is evaluated on several benchmark datasets and is shown to perform comparably to LMNN and ITML.

\paragraph{MDML (Kunapuli \& Shavlik)}
\label{par:mdml}

MDML \citep{Kunapuli2012}, for Mirror Descent Metric Learning, is an attempt of proposing a general framework for online Mahalanobis distance learning. It is based on composite mirror descent \citep{Duchi2010a}, which allows online optimization of many regularized problems. It can accommodate a large class of loss functions and regularizers for which efficient updates are derived, and the algorithm comes with a regret bound. Their study focuses on regularization with the nuclear norm (also called trace norm) introduced by \citet{Fazel2001} and defined as $\|\boldsymbol{M}\|_{*} = \sum_i\sigma_i$, where the $\sigma_i$'s are the singular values of $\boldsymbol{M}$.\footnote{Note that when $\boldsymbol{M}\in \mathbb{S}^{d}_+$, $\|\boldsymbol{M}\|_{*} = \trace(\boldsymbol{M}) = \sum_{i=1}^dM_{ii}$, which is much cheaper to compute.} It is known to be the best convex relaxation of the rank of the matrix and thus nuclear norm regularization tends to induce low-rank matrices.
In practice, MDML has performance comparable to LMNN and ITML, is fast and sometimes induces low-rank solutions, but surprisingly the algorithm was not evaluated on large-scale datasets.

\subsection{Multi-Task Metric Learning}
\label{sec:mt}

This section covers Mahalanobis distance learning for the multi-task setting \citep{Caruana1997}, where given a set of related tasks, one learns a metric for each in a coupled fashion in order to improve the performance on all tasks.

\paragraph{mt-LMNN (Parameswaran \& Weinberger)}
\label{par:mtlmnn}

Multi-Task LMNN\footnote{Source code available at: \url{http://www.cse.wustl.edu/~kilian/code/code.html}} \citep{Parameswaran2010} is a straightforward adaptation of the ideas of Multi-Task SVM \citep{Evgeniou2004} to metric learning. Given $T$ related tasks, they model the problem as learning a shared Mahalanobis metric $d_{\boldsymbol{M}_0}$ as well as task-specific metrics $d_{\boldsymbol{M}_1},\dots,d_{\boldsymbol{M}_t}$ and define the metric for task $t$ as
$$d_t(\boldsymbol{x},\boldsymbol{x'}) = (\boldsymbol{x}-\boldsymbol{x'})^T(\boldsymbol{M}_0+\boldsymbol{M}_t)(\boldsymbol{x}-\boldsymbol{x'}).$$
Note that $\boldsymbol{M}_0+\boldsymbol{M}_t\succeq 0$, hence $d_t$ is a valid pseudo-metric. The LMNN formulation is easily generalized to this multi-task setting so as to learn the metrics jointly, with a specific regularization term defined as follows:
$$\gamma_0\|\boldsymbol{M}_0-\boldsymbol{I}\|^2_\mathcal{F}+\sum_{t=1}^T\gamma_t\|\boldsymbol{M}_t\|^2_\mathcal{F},$$
where $\gamma_t$ controls the regularization of $\boldsymbol{M}_t$. When $\gamma_0\rightarrow\infty$, the shared metric $d_{\boldsymbol{M}_0}$ is simply the Euclidean distance, and the formulation reduces to $T$ independent LMNN formulations. On the other hand, when $\gamma_{t>0}\rightarrow\infty$, the task-specific matrices are simply zero matrices and the formulation reduces to LMNN on the union of all data. In-between these extreme cases, these parameters can be used to adjust the relative importance of each metric: $\gamma_0$ to set the overall level of shared information, and $\gamma_t$ to set the importance of $\boldsymbol{M}_t$ with respect to the shared metric. The formulation remains convex and can be solved using the same efficient solver as LMNN. In the multi-task setting, mt-LMNN clearly outperforms single-task metric learning methods and other multi-task classification techniques such as mt-SVM.

\paragraph{MLCS (Yang et al.)}
\label{par:mlcs}

MLCS \citep{Yang2011} is a different approach to the problem of multi-task metric learning. For each task $t\in\{1,\dots,T\}$, the authors consider learning a Mahalanobis metric
$$d^2_{\boldsymbol{L}_t^T\boldsymbol{L}_t}(\boldsymbol{x},\boldsymbol{x'}) = (\boldsymbol{x}-\boldsymbol{x'})^T\boldsymbol{L}_t^T\boldsymbol{L}_t(\boldsymbol{x}-\boldsymbol{x'}) = (\boldsymbol{L}_t\boldsymbol{x}-\boldsymbol{L}_t\boldsymbol{x'})^T(\boldsymbol{L}_t\boldsymbol{x}-\boldsymbol{L}_t\boldsymbol{x'})$$
parameterized by the transformation matrix $\boldsymbol{L}_t\in\mathbb{R}^{r\times d}$.
They show that $\boldsymbol{L}_t$ can be decomposed into a ``subspace'' part $\boldsymbol{L}_0^t\in\mathbb{R}^{r\times d}$ and a ``low-dimensional metric'' part $\boldsymbol{R}_t\in\mathbb{R}^{r\times r}$ such that $\boldsymbol{L}_t = \boldsymbol{R}_t\boldsymbol{L}^t_0$. The main assumption of MLCS is that all tasks share a common subspace, i.e., $\forall t$, $\boldsymbol{L}^t_0 = \boldsymbol{L}_0$. This parameterization can be used to extend most of metric learning methods to the multi-task setting, although it breaks the convexity of the formulation and is thus subject to local optima. However, as opposed to mt-LMNN, it can be made low-rank by setting $r<d$ and thus has many less parameters to learn. In their work, MLCS is applied to the version of LMNN solved with respect to the transformation matrix \citep{Torresani2006}. The resulting method is evaluated on problems with very scarce training data and study the performance for different values of $r$. It is shown to outperform mt-LMNN, but the setup is a bit unfair to mt-LMNN since it is forced to be low-rank by eigenvalue thresholding.

\paragraph{GPML (Yang et al.)}
\label{par:gpml}

The work of \citet{Yang2012} identifies two drawbacks of previous multi-task metric learning approaches: (i) MLCS's assumption of common subspace is sometimes too strict and leads to a nonconvex formulation, and (ii) the Frobenius regularization of mt-LMNN does not preserve geometry. This property is defined as being the ability to propagate side-information: the task-specific metrics should be regularized so as to preserve the relative distance between training pairs. They introduce the following formulation, which extends any metric learning algorithm to the multi-task setting:
\begin{equation}
\label{eq:gpml}
\begin{aligned}
 \min_{\boldsymbol{M}_0,\dots,\boldsymbol{M}_t\in \mathbb{S}^{d}_+} &&& \sum_{i=1}^t\left(\ell(\boldsymbol{M}_t,\mathcal{S}_t,\mathcal{D}_t,\mathcal{R}_t)+\gamma d_\varphi(\boldsymbol{M}_t,\boldsymbol{M}_0)\right)\quad+\quad\gamma_0d_\varphi(\boldsymbol{A}_0,\boldsymbol{M}_0),
\end{aligned}
\end{equation}
where $\ell(\boldsymbol{M}_t,\mathcal{S}_t,\mathcal{D}_t,\mathcal{R}_t)$ is the loss function for the task $t$ based on the training pairs/triplets (depending on the chosen algorithm), $d_\varphi(\boldsymbol{A},\boldsymbol{B}) = \varphi(\boldsymbol{A}) - \varphi(\boldsymbol{B}) - \trace\left((\nabla\varphi\boldsymbol{B})^T(\boldsymbol{A}-\boldsymbol{B})\right)$ is a Bregman matrix divergence \citep{Dhillon2007} and $\boldsymbol{A}_0$ is a predefined metric (e.g., the identity matrix $\boldsymbol{I}$). mt-LMNN can essentially be recovered from \eqref{eq:gpml} by setting $\varphi(\boldsymbol{A}) = \|\boldsymbol{A}\|^2_\mathcal{F}$ and additional constraints $\boldsymbol{M}_t\succeq\boldsymbol{M}_0$. The authors focus on the von Neumann divergence:
$$d_{VN}(\boldsymbol{A},\boldsymbol{B}) = \trace(\boldsymbol{A}\log\boldsymbol{A}-\boldsymbol{A}\log\boldsymbol{B} - \boldsymbol{A} + \boldsymbol{B}),$$
where $\log\boldsymbol{A}$ is the matrix logarithm of $\boldsymbol{A}$. Like the LogDet divergence mentioned earlier in this survey (\sref{sec:itml}), the von Neumann divergence is known to be rank-preserving and to provide automatic enforcement of positive-semidefiniteness. The authors further show that minimizing this divergence encourages geometry preservation between the learned metrics. Problem \eqref{eq:gpml} remains convex as long as the original algorithm used for solving each task is convex, and can be solved efficiently using gradient descent methods. In the experiments, the method is adapted to LMNN and outperforms single-task LMNN as well as mt-LMNN, especially when training data is very scarce.

\paragraph{TML (Zhang \& Yeung)}
\label{par:tml}

\citet{Zhang2010b} propose a transfer metric learning (TML) approach.\footnote{Source code available at: \url{http://www.cse.ust.hk/~dyyeung/}} They assume that we are given $S$ independent source tasks with enough labeled data and that a Mahalanobis distance $\boldsymbol{M}_s$ has been learned for each task $s$. The goal is to leverage the information of the source metrics to learn a distance $\boldsymbol{M}_t$ for a target task, for which we only have a scarce amount $n_t$ of labeled data. No assumption is made about the relation between the source tasks and the target task: they may be positively/negatively correlated or uncorrelated. The problem is formulated as follows:
\begin{equation}
\label{eq:tml}
\begin{aligned}
\displaystyle\min_{\boldsymbol{M}_t\in \mathbb{S}^{d}_+,\boldsymbol{\Omega} \succeq 0} &&& \frac{2}{n_t^2}\displaystyle\sum_{i<j}\ell\left(y_iy_j\left[1-d_{\boldsymbol{M}_t}^2(\boldsymbol{x}_i,\boldsymbol{x}_j)\right]\right) + \frac{\lambda_1}{2}\|\boldsymbol{M}_t\|_\mathcal{F}^2+ \frac{\lambda_2}{2}\trace(\boldsymbol{\tilde{M}}\boldsymbol{\Omega}^{-1}\boldsymbol{\tilde{M}}^T)\\
 \text{s.t.} &&& \trace(\boldsymbol{\Omega}) = 1,
\end{aligned}
\end{equation}
where $\ell(t) = \max(0,1-t)$ is the hinge loss, $\boldsymbol{\tilde{M}} = \left(\vectorize(\boldsymbol{M}_1),\dots,\vectorize(\boldsymbol{M}_s),\vectorize(\boldsymbol{M}_t)\right)$. The first two terms are classic (loss on all possible pairs and Frobenius regularization) while the third one models the relation between tasks based on a positive definite covariance matrix $\boldsymbol{\Omega}$. Assuming that the source tasks are independent and of equal importance, $\boldsymbol{\Omega}$ can be expressed as
$$\boldsymbol{\Omega} =
\left(\begin{array}{cc}
\alpha\boldsymbol{I}^{(m-1)\times(m-1)} & \boldsymbol{\omega}_m\\
\boldsymbol{\omega}_m & \omega
\end{array}\right),$$
where $\boldsymbol{\omega}_m$ denotes the task covariances between the target task and the source tasks, and $\omega$ denotes the variance of the target task. Problem \eqref{eq:tml} is convex and is solved using an alternating procedure that is guaranteed to converge to the global optimum: (i) fixing $\boldsymbol{\Omega}$ and solving for $\boldsymbol{M}_t$, which is done online with an algorithm similar to RDML, and (ii) fixing $\boldsymbol{M}_t$ and solving for $\boldsymbol{\Omega}$, leading to a second-order cone program whose number of variables and constraints is linear in the number of tasks. In practice, TML consistently outperforms metric learning methods without transfer when training data is scarce.

%[METTRE CETTE APPROCHE AVEC MULTI TASK METRIC LEARNING?]

\subsection{Other Approaches}
\label{sec:othermaha}

In this section, we describe a few approaches that are outside the scope of the previous categories. The first two (LPML and SML) fall into the category of sparse metric learning methods. BoostMetric is inspired from the theory of boosting. DML-$p$ revisits the original metric learning formulation of Xing et al. RML deals with the presence of noisy constraints. Finally, MLR learns a metric for solving a ranking task.

%[METTRE LES DEUX PREMIERS DANS UNE CATEGORIE SPARSE METRIC LEARNING?]

\paragraph{LPML (Rosales \& Fung)}
\label{par:lpml}

The method of \citet{Rosales2006} aims at learning matrices with entire columns/rows set to zero, thus making $\boldsymbol{M}$ low-rank. For this purpose, they use $L_1$ norm regularization and, restricting their framework to diagonal dominant matrices, they are able to formulate the problem as a linear program that can be solved efficiently. However, $L_1$ norm regularization favors sparsity at the entry level only, not specifically at the row/column level, even though in practice the learned matrix is sometimes low-rank. Furthermore, the approach is less general than Mahalanobis distances due to the restriction to diagonal dominant matrices.

\paragraph{SML (Ying et al.)}
\label{par:sml}

SML\footnote{Source code is not available but is indicated as ``coming soon'' by the authors. Check:\\ \url{http://www.enm.bris.ac.uk/staff/xyy/software.html}} \citep{Ying2009}, for Sparse Metric Learning, is a distance learning approach that regularizes $\boldsymbol{M}$ with the mixed $L_{2,1}$ norm defined as
$$\|\boldsymbol{M}\|_{2,1} = \sum_{i=1}^d \|\boldsymbol{M}_i\|_2,$$
which tends to zero out entire rows of $\boldsymbol{M}$ (as opposed to the $L_1$ norm used in LPML), and therefore performs feature selection. More precisely, they set $\boldsymbol{M} = \boldsymbol{U}^T\boldsymbol{W}\boldsymbol{U}$, where $\boldsymbol{U}\in\mathbb{O}^d$ (the set of $d\times d$ orthonormal matrices) and $\boldsymbol{W}\in\mathbb{S}^{d}_+$, and solve the following problem:
\begin{equation}
\label{eq:sml}
\begin{aligned}
 \min_{\boldsymbol{U}\in\mathbb{O}^d,\boldsymbol{W}\in \mathbb{S}^{d}_+} &&& \|\boldsymbol{W}\|_{2,1}\quad+\quad\gamma\sum_{i,j,k}\xi_{ijk}\\
 \text{s.t.} &&& d_{\boldsymbol{M}}^2(\boldsymbol{x}_i,\boldsymbol{x}_k) - d_{\boldsymbol{M}}^2(\boldsymbol{x}_i,\boldsymbol{x}_j) \geq 1 - \xi_{ijk} && \forall (\boldsymbol{x}_i,\boldsymbol{x}_j,\boldsymbol{x}_k)\in \mathcal{R},
\end{aligned}
\end{equation}
where $\gamma\geq 0$ is the trade-off parameter. Unfortunately, $L_{2,1}$ regularized problems are typically difficult to optimize. Problem \eqref{eq:sml} is reformulated as a min-max problem and solved using smooth optimization \citep{Nesterov2005}. Overall, the algorithm has a fast convergence rate but each iteration has an $O(d^3)$ complexity. The method performs well in practice while achieving better dimensionality reduction than full-rank methods such as \citet{Rosales2006}. However, it cannot be applied to high-dimensional problems due to the complexity of the algorithm. Note that the same authors proposed a unified framework for sparse metric learning \citep{Huang2009,Huang2011}.

\paragraph{BoostMetric (Shen et al.)}
\label{par:boost}

BoostMetric\footnote{Source code available at: \url{http://code.google.com/p/boosting/}} \citep{Shen2009,Shen2012} adapts to Mahalanobis distance learning the ideas of boosting, where a good hypothesis is obtained through a weighted combination of so-called ``weak learners'' \citep[see the recent book on this matter by][]{Schapire2012}. The method is based on the property that any PSD matrix can be decomposed into a positive linear combination of trace-one rank-one matrices. This kind of matrices is thus used as weak learner and the authors adapt the popular boosting algorithm Adaboost \citep{Freund1995} to this setting. The resulting algorithm is quite efficient since it does not require full eigenvalue decomposition but only the computation of the largest eigenvalue. In practice, BoostMetric achieves competitive performance but typically requires a very large number of iterations for high-dimensional datasets. \citet{Bi2011} further improve the scalability of the approach, while \cite{Liu2012} introduce regularization on the weights as well as a term to reduce redundancy among the weak learners.

\paragraph{DML-$p$ (Ying et al., Cao et al.)}
\label{par:dmlp}

The work of \citet{Ying2012,Cao2012} revisit MMC, the original approach of \citet{Xing2002}, by investigating the following formulation, called DML-$p$:
\begin{equation}
\label{eq:dml-p}
\begin{aligned}
\max_{\boldsymbol{M}\in \mathbb{S}^{d}_+} &&& \left(\frac{1}{|\mathcal{D}|}\displaystyle\sum_{(\boldsymbol{x}_i,\boldsymbol{x}_j)\in \mathcal{D}}[d_{\boldsymbol{M}}(\boldsymbol{x}_i,\boldsymbol{x}_j)]^{2p}\right)^{1/p}\\
\text{s.t.} &&& \displaystyle\sum_{(\boldsymbol{x}_i,\boldsymbol{x}_j)\in \mathcal{S}}d_{\boldsymbol{M}}^2(\boldsymbol{x}_i,\boldsymbol{x}_j) \leq 1.
\end{aligned}
\end{equation}
Note that by setting $p=0.5$ we recover MMC. The authors show that  \eqref{eq:dml-p} is convex for $p\in(-\infty,1)$ and can be cast as a well-known eigenvalue optimization problem called ``minimizing the maximal eigenvalue of a symmetric matrix''. They further show that it can be solved efficiently using a first-order algorithm that only requires the computation of the largest eigenvalue at each iteration (instead of the costly full eigen-decomposition used by Xing et al.). Experiments show competitive results and low computational complexity. A general drawback of DML-$p$ is that it is not clear how to accommodate a regularizer (e.g., sparse or low-rank).

\paragraph{RML (Huang et al.)}
\label{par:rml}

Robust Metric Learning \citep{Huang2010} is a method that can successfully deal with the presence of noisy/incorrect training constraints, a situation that can arise when they are not derived from class labels but from side information such as users' implicit feedback. The approach is based on robust optimization \citep{Ben-Tal2009}: assuming that a proportion $1-\eta$ of the $m$ training constraints (say triplets) are incorrect, it minimizes some loss function $\ell$ for any $\eta$ fraction of the triplets:
\begin{equation}
\label{eq:rml}
\begin{aligned}
 \min_{\boldsymbol{M}\in \mathbb{S}^{d}_+,t} &&& t\quad+\quad\frac{\lambda}{2}\|\boldsymbol{M}\|_{\mathcal{F}}\\
 \text{s.t.} &&& t \geq \sum_{i=1}^mq_{i}\ell\left(d^2_{\boldsymbol{M}}(\boldsymbol{x}_i,\boldsymbol{x''_i})-d^2_{\boldsymbol{M}}(\boldsymbol{x}_i,\boldsymbol{x'_i})\right), && \forall\boldsymbol{q}\in\mathcal{Q}(\eta),
\end{aligned}
\end{equation}
where $\ell$ is taken to be the hinge loss and $\mathcal{Q}(\eta)$ is defined as
$$\mathcal{Q}(\eta) = \left\{\boldsymbol{q}\in\{0,1\}^m:\sum_{i=1}^mq_i\leq \eta m\right\}.$$
In other words, Problem \eqref{eq:rml} minimizes the worst-case violation over all possible sets of correct constraints.
$\mathcal{Q}(\eta)$ can be replaced by its convex hull, leading to a semi-definite program with an infinite number of constraints. This can be further simplified into a convex minimization problem that can be solved either using subgradient descent or smooth optimization \citep{Nesterov2005}. However, both of these require a projection onto the PSD cone. Experiments on standard datasets show good robustness for up to 30\% of incorrect triplets, while the performance of other methods such as LMNN is greatly damaged.

\paragraph{MLR (McFee \& Lankriet)}
\label{par:mlr}

The idea of MLR \citep{McFee2010}, for Metric Learning to Rank, is to learn a metric for a ranking task, where given a query instance, one aims at producing a ranked list of examples where relevant ones are ranked higher than irrelevant ones.\footnote{Source code is available at: \url{http://www-cse.ucsd.edu/~bmcfee/code/mlr}} Let $\mathcal{P}$ the set of all permutations (i.e., possible rankings) over the training set. Given a Mahalanobis distance $d_{\boldsymbol{M}}^2$ and a query $\boldsymbol{x}$, the predicted ranking $p\in\mathcal{P}$ consists in sorting the instances by ascending $d_{\boldsymbol{M}}^2(\boldsymbol{x},\cdot)$.
The metric learning $\boldsymbol{M}$ is based on Structural SVM \citep{Tsochantaridis2005}:
\begin{equation}
\label{eq:mlr}
\begin{aligned}
 \min_{\boldsymbol{M}\in \mathbb{S}^{d}_+} &&& \|\boldsymbol{M}\|_{*}\quad+\quad C\sum_i\xi_i\\
 \text{s.t.} &&& \innerp{\boldsymbol{M},\psi(\boldsymbol{x}_i,p_i)-\psi(\boldsymbol{x}_i,p)}_\mathcal{F} \geq \Delta(p_i,p) - \xi_i && \forall i\in\{1,\dots,n\},p\in\mathcal{P},
\end{aligned}
\end{equation}
where $\|\boldsymbol{M}\|_{*} = \trace(\boldsymbol{M})$ is the nuclear norm, $C\geq 0$ the trade-off parameter, $\innerp{\boldsymbol{A},\boldsymbol{B}}_\mathcal{F}=\sum_{i,j}A_{ij}B_{ij}$ the Frobenius inner product, $\psi:\mathbb{R^d}\times \mathcal{P}\rightarrow \mathbb{S}^d$ the feature encoding of an input-output pair $(\boldsymbol{x}_i,p)$,\footnote{The feature map $\psi$ is designed such that the ranking $p$ which maximizes $\innerp{\boldsymbol{M},\psi(\boldsymbol{x},p)}_\mathcal{F}$ is the one given by ascending $d_{\boldsymbol{M}}^2(\boldsymbol{x},\cdot)$.} and $\Delta(p_i,p)\in[0,1]$ the ``margin'' representing the loss of predicting ranking $p$ instead of the true ranking $p_i$. In other words, $\Delta(p_i,p)$ assesses the quality of ranking $p$ with respect to the best ranking $p_i$ and can be evaluated using several measures, such as the Area Under the ROC Curve (AUC), Precision-at-$k$ or Mean Average Precision (MAP). Since the number of constraints is super-exponential in the number of training instances, the authors solve \eqref{eq:mlr} using a 1-slack cutting-plane approach \citep{Joachims2009a} which essentially iteratively optimizes over a small set of active constraints (adding the most violated ones at each step) using subgradient descent. However, the algorithm requires a full eigendecomposition of $\boldsymbol{M}$ at each iteration, thus MLR does not scale well with the dimensionality of the data. In practice, it is competitive with other metric learning algorithms for $k$-NN classification and a structural SVM algorithm for ranking, and can induce low-rank solutions due to the nuclear norm. \citet{Lim2013} propose R-MLR, an extension to MLR to deal with the presence of noisy features\footnote{Notice that this is different from noisy side information, which was investigated by the method RML \citep{Huang2010} presented earlier in this section.} using the mixed $L_{2,1}$ norm as in SML \citep{Ying2009}. R-MLR is shown to be able to ignore most of the irrelevant features and outperforms MLR in this situation.

%% file: other_numerical.tex
\section{Other Advances in Metric Learning}
\label{sec:recentml}

So far, we focused on (linear) Mahalanobis metric learning which has inspired a large amount of work during the past ten years. In this section, we cover other advances and trends in metric learning for feature vectors. Most of the section is devoted to (fully and weakly) supervised methods. In \sref{sec:otherlin}, we address linear similarity learning. \sref{sec:nonlinear} deals with nonlinear metric learning (including the kernelization of linear methods), \sref{sec:local} with local metric learning and \sref{sec:histogram} with metric learning for histogram data. \sref{sec:genml} presents the recently-developed frameworks for deriving generalization guarantees for supervised metric learning. We conclude this section with a review of semi-supervised metric learning (\sref{sec:semi}).

\subsection{Linear Similarity Learning}
\label{sec:otherlin}

Although most of the work in linear metric learning has focused on the Mahalanobis distance, other linear measures, in the form of similarity functions,  have recently attracted some interest. These approaches are often motivated by the perspective of more scalable algorithms due to the absence of PSD constraint.

\paragraph{SiLA (Qamar et al.)}
\label{par:sila}

SiLA \citep{Qamar2008} is an approach for learning similarity functions of the following form:
$$K_{\boldsymbol{M}}(\boldsymbol{x},\boldsymbol{x}')=\frac{\boldsymbol{x}^T\boldsymbol{M}\boldsymbol{x'}}{N(\boldsymbol{x},\boldsymbol{x'})},$$
where $\boldsymbol{M}\in\mathbb{R}^{d\times d}$ is not required to be PSD nor symmetric, and $N(\boldsymbol{x},\boldsymbol{x'})$ is a normalization term which depends on $\boldsymbol{x}$ and $\boldsymbol{x'}$. This similarity function can be seen as a generalization of the cosine similarity, widely used in text and image retrieval \citep[see for instance][]{Baeza-Yates1999,Sivic2009}. The authors build on the same idea of ``target neighbors'' that was introduced in LMNN, but optimize the similarity in an online manner with an algorithm based on voted perceptron. At each step, the algorithm goes through the training set, updating the matrix when an example does not satisfy a criterion of separation. The authors present theoretical results that follow from the voted perceptron theory in the form of regret bounds for the separable and inseparable cases. %SiLA is compared to Mahalanobis metric learning approaches on three datasets. It seems to perform fine but has a rather slow convergence rate and may suffer from its lack of regularization.
In subsequent work, \citet{Qamar2012} study the relationship between SiLA and RELIEF, an online feature reweighting algorithm.

\paragraph{gCosLA (Qamar \& Gaussier)}
\label{par:gcosla}

gCosLA \citep{Qamar2009} learns generalized cosine similarities of the form
$$K_{\boldsymbol{M}}(\boldsymbol{x},\boldsymbol{x}')=\frac{\boldsymbol{x}^T\boldsymbol{M}\boldsymbol{x'}}{\sqrt{\boldsymbol{x}^T\boldsymbol{M}\boldsymbol{x}}\sqrt{\boldsymbol{x'}^T\boldsymbol{M}\boldsymbol{x'}}},$$
where $\boldsymbol{M}\in\mathbb{S}_+^d$. It corresponds to a cosine similarity in the projection space implied by $\boldsymbol{M}$. The algorithm itself, an online procedure, is very similar to that of POLA (presented in \sref{sec:onlineml}). Indeed, they essentially use the same loss function and also have a two-step approach: a projection onto the set of arbitrary matrices that achieve zero loss on the current example pair, followed by a projection back onto the PSD cone. The first projection is different from POLA (since the generalized cosine has a normalization factor that depends on $\boldsymbol{M}$) but the authors manage to derive a closed-form solution. The second projection is based on a full eigenvalue decomposition of $\boldsymbol{M}$, making the approach costly as dimensionality grows. A regret bound for the algorithm is provided and it is shown experimentally that gCosLA converges in fewer iterations than SiLA and is generally more accurate. Its performance is competitive with LMNN and ITML. Note that \citet{Nguyen2010} optimize the same form of similarity based on a nonconvex formulation.

\paragraph{OASIS (Chechik et al.)}
\label{par:oasis}

OASIS\footnote{Source code available at: \url{http://ai.stanford.edu/~gal/Research/OASIS/}} \citep{Chechik2009,Chechik2010} learns a bilinear similarity with a focus on large-scale problems. The bilinear similarity has been used for instance in image retrieval \citep{Deng2011} and has the following simple form:
$$K_{\boldsymbol{M}}(\boldsymbol{x},\boldsymbol{x'})Â = \boldsymbol{x}^T\boldsymbol{M}\boldsymbol{x'},$$
where $\boldsymbol{M} \in \mathbb{R}^{d\times d}$ is not required to be PSD nor symmetric. In other words, it is related to the (generalized) cosine similarity but does not include normalization nor PSD constraint.
Note that when $\boldsymbol{M}$ is the identity matrix, $K_{\boldsymbol{M}}$ amounts to an unnormalized cosine similarity.
The bilinear similarity has two advantages. First, it is efficiently computable for sparse inputs: if $\boldsymbol{x}$ and $\boldsymbol{x'}$ have $k_1$ and $k_2$ nonzero features, $K_{\boldsymbol{M}}(\boldsymbol{x},\boldsymbol{x'})$ can be computed in $O(k_1k_2)$ time. Second, unlike the Mahalanobis distance, it can define a similarity measure between instances of different dimension (for example, a document and a query) if a rectangular matrix $\boldsymbol{M}$ is used.
Since $\boldsymbol{M}\in\mathbb{R}^{d\times d}$ is not required to be PSD, Chechik et al. are able to optimize $K_{\boldsymbol{M}}$ in an online manner using a simple and efficient algorithm, which belongs to the family of Passive-Aggressive algorithms \citep{Crammer2006}. The initialization is $\boldsymbol{M}=\boldsymbol{I}$, then at each step $t$, the algorithm draws a triplet $(\boldsymbol{x}_i,\boldsymbol{x}_j,\boldsymbol{x}_k)\in \mathcal{R}$ and solves the following convex problem:
\begin{equation}
\label{eq:oasis}
\begin{aligned}
\boldsymbol{M}^t & = & \displaystyle\argmin_{\boldsymbol{M},\xi} &&& \frac{1}{2}\|\boldsymbol{M}-\boldsymbol{M}^{t-1}\|_{\mathcal{F}}^2+C\xi\\
&& \text{s.t.} &&& 1-d^2_{\boldsymbol{M}}(\boldsymbol{x}_i,\boldsymbol{x}_j)+d^2_{\boldsymbol{M}}(\boldsymbol{x}_i,\boldsymbol{x}_k)\leq \xi\\
&& &&& \xi \geq 0,
\end{aligned}
\end{equation}
where $C\geq 0$ is the trade-off parameter between minimizing the loss and staying close from the matrix obtained at the previous step. Clearly, if $1-d^2_{\boldsymbol{M}}(\boldsymbol{x}_i,\boldsymbol{x}_j)+d^2_{\boldsymbol{M}}(\boldsymbol{x}_i,\boldsymbol{x}_k) \leq 0$, then $\boldsymbol{M}^t=\boldsymbol{M}^{t-1}$ is the solution of \eqref{eq:oasis}. Otherwise, the solution is obtained from a simple closed-form update. In practice, OASIS achieves competitive results on medium-scale problems and unlike most other methods, is scalable to problems with millions of training instances. However, it cannot incorporate complex regularizers.
Note that the same authors derived two more algorithms for learning bilinear similarities as applications of more general frameworks. The first one is based on online learning in the manifold of low-rank matrices \citep{Shalit2010,Shalit2012} and the second on adaptive regularization of weight matrices \citep{Crammer2012}.

\paragraph{SLLC (Bellet et al.)}
\label{par:sllc}

Similarity Learning for Linear Classification \citep{Bellet2012a} takes an original angle by focusing on metric learning for linear classification. As opposed to pair and triplet-based constraints used in other approaches, the metric is optimized to be $(\epsilon,\gamma,\tau)$-good \citep{Balcan2008a}, a property based on an average over some points which has a deep connection with the performance of a sparse linear classifier built from such a similarity.
SLLC learns a bilinear similarity $K_{\boldsymbol{M}}$ and is formulated as an efficient unconstrained quadratic program:
\begin{eqnarray}
\label{eq:sllc}
\displaystyle\min_{\boldsymbol{M}\in\mathbb{R}^{d\times d}} & \frac{1}{n}\displaystyle\sum_{i=1}^{n}\ell(1 - y_i\frac{1}{\gamma |\mathcal{R}|}\sum_{\boldsymbol{x}_j\in\mathcal{R}}y_jK_{\boldsymbol{M}}(\boldsymbol{x}_i,\boldsymbol{x}_j)) \quad+\quad\beta\|\boldsymbol{M}\|_{\mathcal{F}}^2,
\end{eqnarray}
where $\mathcal{R}$ is a set of reference points randomly selected from the training sample, $\gamma$ is the margin parameter, $\ell$ is the hinge loss and $\beta$ the regularization parameter. Problem \eqref{eq:sllc} essentially learns $K_{\boldsymbol{M}}$ such that training examples are more similar on average to reference points of the same class than to reference points of the opposite class by a margin $\gamma$. In practice, SLLC is competitive with traditional metric learning methods, with the additional advantage of inducing extremely sparse classifiers. A drawback of the approach is that linear classifiers (unlike $k$-NN) cannot naturally deal with the multi-class setting, and thus one-vs-all or one-vs-one strategies must be used.

\paragraph{RSL (Cheng)}
\label{par:rsl}

As OASIS and SLLC, \citet{Cheng2013} also proposes to learn a bilinear similarity, but focuses on the setting of pair matching (predicting whether two pairs are similar). Pairs are of the form $(\boldsymbol{x},\boldsymbol{x'})$, where $\boldsymbol{x}\in \mathbb{R}^d$ and $\boldsymbol{x'}\in \mathbb{R}^{d'}$ potentially have different dimensionality, thus one has to learn a rectangular matrix $\boldsymbol{M}\in\mathbb{R}^{d\times d'}$. This is a relevant setting for matching instances from different domains, such as images with different resolutions, or queries and documents. The matrix $\boldsymbol{M}$ is set to have fixed rank $r \ll\min(d,d')$. RSL (Riemannian Similarity Learning) is formulated as follows:
\begin{equation}
\label{eq:rsl}
\begin{aligned}
\max_{\boldsymbol{M}\in \mathbb{R}^{d\times d'}} &&& \displaystyle\sum_{(\boldsymbol{x}_i,\boldsymbol{x}_j)\in \mathcal{S}\cup\mathcal{D}}\ell(\boldsymbol{x}_i,\boldsymbol{x}_j,y_{ij})\quad+\quad\|\boldsymbol{M}\|_\mathcal{F}\\
\text{s.t.} &&& \rank (\boldsymbol{M}) = r,
\end{aligned}
\end{equation}
where $\ell$ is some differentiable loss function (such as the log loss or the squared hinge loss). The optimization is carried out efficiently using recent advances in optimization over Riemannian manifolds \citep{Absil2008} and based on the low-rank factorization of $\boldsymbol{M}$. At each iteration, the procedure finds a descent direction in the tangent space of the current solution, and a retractation step to project the obtained matrix back to the low-rank manifold. It outputs a local minimum of \eqref{eq:rsl}. Experiments are conducted on pair-matching problems where RSL achieves state-of-the-art results using a small rank matrix.

\subsection{Nonlinear Methods}
\label{sec:nonlinear}

As we have seen, work in supervised metric learning has focused on linear metrics because they are more convenient to optimize (in particular, it is easier to derive convex formulations with the guarantee of finding the global optimum) and less prone to overfitting.
In some cases, however, there is nonlinear structure in the data that linear metrics are unable to capture.
The kernelization of linear methods can be seen as a satisfactory solution to this problem. This strategy is explained in \sref{sec:kernelization}. The few approaches consisting in directly learning nonlinear forms of metrics are addressed in \sref{sec:nonlineardirect}.

\subsubsection{Kernelization of Linear Methods}
\label{sec:kernelization}

The idea of kernelization is to learn a linear metric in the nonlinear feature space induced by a kernel function and thereby combine the best of both worlds, in the spirit of what is done in SVM. Some metric learning approaches have been shown to be kernelizable \citep[see for instance][]{Schultz2003,Shalev-Shwartz2004,Hoi2006,Torresani2006,Davis2007} using specific arguments, but in general kernelizing a particular metric algorithm is not trivial: a new formulation of the problem has to be derived, where interface to the data is limited to inner products, and sometimes a different implementation is necessary. Moreover, when kernelization is possible, one must learn a $n\times n$ matrix. As the number of training examples $n$ gets large, the problem becomes intractable.

Recently though, several authors \citep{Chatpatanasiri2010,Zhang2010a} have proposed general kernelization methods based on Kernel Principal Component Analysis \citep{Scholkopf1998}, a nonlinear extension of PCA \citep{Pearson1901}. In short, KPCA implicitly projects the data into the nonlinear (potentially infinite-dimensional) feature space induced by a kernel and performs dimensionality reduction in that space. The (unchanged) metric learning algorithm can then be used to learn a metric in that nonlinear space---this is referred to as the ``KPCA trick''. Chatpatanasiri et al. \citeyearpar{Chatpatanasiri2010} showed that the KPCA trick is theoretically sound for unconstrained metric learning algorithms (they prove representer theorems). Another trick (similar in spirit in the sense that it involves some nonlinear preprocessing of the feature space) is based on kernel density estimation and allows one to deal with both numerical and categorical attributes \citep{He2013}.
General kernelization results can also be obtained from the equivalence between Mahalanobis distance learning in kernel space and linear transformation kernel learning \citep{Jain2010,Jain2012}, but are restricted to spectral regularizers. Lastly, \citet{Wang2011} address the problem of choosing an appropriate kernel function by proposing a multiple kernel framework for metric learning.

Note that kernelizing a metric learning algorithm may drastically improve the quality of the learned metric on highly nonlinear problems, but may also favor overfitting (because pair or triplet-based constraints become much easier to satisfy in a nonlinear, high-dimensional kernel space) and thereby lead to poor generalization performance.

\subsubsection{Learning Nonlinear Forms of Metrics}
\label{sec:nonlineardirect}

A few approaches have tackled the direct optimization of nonlinear forms of metrics. These approaches are subject to local optima and more inclined to overfit the data, but have the potential to significantly outperform linear methods on some problems.
%We first describe methods for classic numerical data, and then deal with approaches that focus on learning nonlinear metrics for histogram data (i.e., instances lying on the probability simplex).

\paragraph{LSMD (Chopra et al.)}
\label{par:lsmd}

\citet{Chopra2005} pioneered the nonlinear metric learning literature. They learn a nonlinear projection $G_W(\boldsymbol{x})$ parameterized by a vector $W$ such that the $L_1$ distance in the low-dimensional target space $\|G_W(\boldsymbol{x})-G_W(\boldsymbol{x'})\|_1$ is small for positive pairs and large for negative pairs. No assumption is made about the nature of $G_W$: the parameter $W$ corresponds to the weights in a convolutional neural network and can thus be an arbitrarily complex nonlinear mapping. These weights are learned through back-propagation and stochastic gradient descent so as to minimize a loss function designed to make the distance for positive pairs smaller than the distance of negative pairs by a given margin. Due to the use of neural networks, the approach suffers from local optimality and needs careful tuning of the many hyperparameters, requiring a significant amount of validation data in order to avoid overfitting. This leads to a high computational complexity. Nevertheless, the authors demonstrate the usefulness of LSMD on face verification tasks.

\paragraph{NNCA (Salakhutdinov \& Hinton)}
\label{par:nnca}

Nonlinear NCA \citep{Salakhutdinov2007} is another distance learning approach based on deep learning. NNCA first learns a nonlinear, low-dimensional representation of the data using a deep belief network (stacked Restricted Boltzmann Machines) that is pretrained layer-by-layer in an unsupervised way. In a second step, the parameters of the last layer are fine-tuned by optimizing the NCA objective (\sref{sec:knnml}). Additional unlabeled data can be used as a regularizer by minimizing their reconstruction error. Although it suffers from the same limitations as LSMD due to its deep structure, NNCA is shown to perform well when enough data is available. For instance, on a digit recognition dataset, NNCA based on a 30-dimensional nonlinear representation significantly outperforms $k$-NN in the original pixel space as well as NCA based on a linear space of same dimension.

\paragraph{SVML (Xu et al.)}
\label{par:svml}

\citet{Xu2012b} observe that learning a Mahalanobis distance with an existing algorithm and plugging it into a RBF kernel does not significantly improve SVM classification performance. They instead propose Support Vector Metric Learning (SVML), an algorithm that alternates between (i) learning the SVM model with respect to the current Mahalanobis distance and (ii) learning a Mahalanobis distance that minimizes a surrogate of the validation error of the current SVM model. Since the latter step is nonconvex in any event (due to the nonconvex loss function), the authors optimize the distance based on the decomposition $\boldsymbol{L}^T\boldsymbol{L}$, thus there is no PSD constraint and the approach can be made low-rank. Frobenius regularization on $\boldsymbol{L}$ may be used to avoid overfitting. The optimization procedure is done using a gradient descent approach and is rather efficient although subject to local minima. Nevertheless, SVML significantly improves standard SVM results.

\paragraph{GB-LMNN (Kedem et al.)}
\label{par:gblmnn}

\citet{Kedem2012} propose Gradient-Boosted LMNN, a nonlinear method consisting in generalizing the Euclidean distance with a nonlinear transformation $\phi$ as follows:
$$d_\phi(\boldsymbol{x},\boldsymbol{x'}) = \|\phi(\boldsymbol{x})-\phi(\boldsymbol{x'})\|_2.$$
This nonlinear mapping takes the form of an additive function $\phi = \phi_0 + \alpha\sum_{t=1}^Th_t$, where $h_1,\dots,h_T$ are gradient boosted regression trees \citep{Friedman2001} of limited depth $p$ and $\phi_0$ corresponds to the mapping learned by linear LMNN. They once again use the same objective function as LMNN and are able to do the optimization efficiently, building on gradient boosting. On an intuitive level, the tree selected by gradient descent at each iteration divides the space into $2^p$ regions, and instances falling in the same region are translated by the same vector---thus examples in different regions are translated in different directions. Dimensionality reduction can be achieved by learning trees with $r$-dimensional output. In practice, GB-LMNN seems quite robust to overfitting and performs well, often achieving comparable or better performance than LMNN and ITML.

\paragraph{HDML (Norouzi et al.)}
\label{par:hdml}

Hamming Distance Metric Learning \citep{Norouzi2012} proposes to learn mappings from real-valued vectors to binary codes on which the Hamming distance performs well.\footnote{Source code available at: \url{https://github.com/norouzi/hdml}} Recall that the Hamming distance $d_H$ between two binary codes of same length is simply the number of bits on which they disagree. A great advantage of working with binary codes is their small storage cost and the fact that exact neighbor search can be done in sublinear time \citep{Norouzi2012a}. The goal here is to optimize a mapping $b(\boldsymbol{x})$ that projects a $d$-dimensional real-valued input $\boldsymbol{x}$ to a $q$-dimensional binary code. The mapping takes the general form:
$$b(\boldsymbol{x};\boldsymbol{w}) = \sign\left(f(\boldsymbol{x};\boldsymbol{w})\right),$$
where $f:\mathbb{R}^d\rightarrow\mathbb{R}^q$ can be any function differentiable in $\boldsymbol{w}$, $\sign(\cdot)$ is the element-wise sign function and $\boldsymbol{w}$ is a real-valued vector representing the parameters to be learned. For instance, $f$ can be a nonlinear transform obtained with a multilayer neural network. Given a relative constraint $(\boldsymbol{x}_i,\boldsymbol{x}_j,\boldsymbol{x}_k)\in \mathcal{R}$, denote by $\boldsymbol{h}_i$, $\boldsymbol{h}_j$ and $\boldsymbol{h}_k$ their corresponding binary codes given by $b$. The loss is then given by
$$\ell(\boldsymbol{h}_i,\boldsymbol{h}_j,\boldsymbol{h}_k) = \left[1 - d_H\left(\boldsymbol{h}_i,\boldsymbol{h}_k\right) + d_H\left(\boldsymbol{h}_i,\boldsymbol{h}_j\right)\right]_+.$$
In the other words, the loss is zero when the Hamming distance between $\boldsymbol{h}_i$ and $\boldsymbol{h}_j$ is a at least one bit smaller than the distance between $\boldsymbol{h}_i$ and $\boldsymbol{h}_k$. HDML is formalized as a loss minimization problem with $L_2$ norm regularization on $\boldsymbol{w}$. This objective function is nonconvex and discontinuous, but the authors propose to optimize a continuous upper bound on the loss which can be computed in $O(q^2)$ time, which is efficient as long as the code length $q$ remains small. In practice, the objective is optimized using a stochastic gradient descent approach. Experiments show that relatively short codes obtained by nonlinear mapping are sufficient to achieve few constraint violations, and that a $k$-NN classifier based on these codes can achieve competitive performance with state-of-the-art classifiers.
\citet{Neyshabur2013} later showed that using asymmetric codes can lead to shorter encodings while maintaining similar performance.

\subsection{Local Metric Learning}
\label{sec:local}

The methods studied so far learn a global (linear or nonlinear) metric. However, if the data is heterogeneous, a single metric may not well capture the complexity of the task and it might be beneficial to use multiple local metrics that vary across the space (e.g., one for each class or for each instance).\footnote{The work of \citet{Frome2007} is one of the first to propose to learn multiple local metrics. However, their approach is specific to computer vision so we chose not to review it here.} This can often be seen as approximating the geodesic distance defined by a metric tensor \citep[see][for a review on this matter]{Ramanan2011}.
It is typically crucial that the local metrics be learned simultaneously in order to make them meaningfully comparable and also to alleviate overfitting. Local metric learning has been shown to significantly outperform global methods on some problems, but typically comes at the expense of higher time and memory requirements. Furthermore, they usually do not give rise to a consistent global metric, although some recent work partially addresses this issue \citep{Zhan2009,Hauberg2012}.

\paragraph{M$^2$-LMNN (Weinberger \& Saul)}
\label{par:mmlmnn}

Multiple Metrics LMNN\footnote{Source code available at: \url{http://www.cse.wustl.edu/~kilian/code/code.html}} \citep{Weinberger2008a,Weinberger2009} learns several Mahalanobis distances in different parts of the space. As a preprocessing step, training data is partitioned in $C$ clusters. These can be obtained either in a supervised way (using class labels) or without supervision (e.g., using $K$-Means). Then, $C$ metrics (one for each cluster) are learned in a coupled fashion in the form of a generalization of the LMNN's objective, where the distance to a target neighbor or an impostor $\boldsymbol{x}$ is measured under the local metric associated with the cluster to which $\boldsymbol{x}$ belongs. In practice, M$^2$-LMNN can yield significant improvements over standard LMNN (especially with supervised clustering), but this comes at the expense of a higher computational cost, and important overfitting (since each local metric can be overly specific to its region) unless a large validation set is used \citep{Wang2012b}.

\paragraph{GLML (Noh et al.)}
\label{par:glml}

The work of \citet{Noh2010}, Generative Local Metric Learning, aims at leveraging the power of generative models (known to outperform purely discriminative models when the training set is small) in the context of metric learning. They focus on nearest neighbor classification and express the expected error of a 1-NN classifier as the sum of two terms: the asymptotic probability of misclassification and a metric-dependent term representing the bias due to finite sampling. They show that this bias can be minimized locally by learning a Mahalanobis distance $d_{\boldsymbol{M_i}}$ at each training point $\boldsymbol{x}_i$. This is done by solving, for each training instance, an independent semidefinite program that has an analytical solution. Each matrix $\boldsymbol{M}_i$ is further regularized towards a diagonal matrix in order to alleviate overfitting. Since each local metric is computed independently, GLML can be very scalable. Its performance is competitive on some datasets (where the assumption of Gaussian distribution to model the distribution of data is reasonable) but can perform very poorly on more complex problems \citep{Wang2012b}. Note that GLML does not straightforwardly extend to the $k$-NN setting for $k > 1$. \citet{Shi2011} use GLML metrics as base kernels to learn a global kernel in a discriminative manner.

\paragraph{Bk-means (Wu et al.)}
\label{par:bkmeans}

\citet{Wu2009,Wu2012} propose to learn Bregman distances (or Bregman divergences), a family of metrics that do not necessarily satisfy the triangle inequality or symmetry \citep{Bregman1967}. Given the strictly convex and twice differentiable function $\varphi:\mathbb{R}^d\rightarrow \mathbb{R}$, the Bregman distance is defined as:
$$d_\varphi(\boldsymbol{x},\boldsymbol{x'}) = \varphi(\boldsymbol{x})-\varphi(\boldsymbol{x'}) - (\boldsymbol{x}-\boldsymbol{x'})^T\nabla\varphi(\boldsymbol{x'}).$$
It generalizes many widely-used measures: the Mahalanobis distance is recovered by setting $\varphi(\boldsymbol{x}) = \frac{1}{2}\boldsymbol{x}^T\boldsymbol{M}\boldsymbol{x}$, the KL divergence \citep{Kullback1951} by choosing $\varphi(\boldsymbol{p}) = \sum_{i=1}^dp_i\log p_i$ (here, $\boldsymbol{p}$ is a discrete probability distribution), etc. Wu et al. consider the following symmetrized version:
\begin{eqnarray*}
d_\varphi(\boldsymbol{x},\boldsymbol{x'}) & = & \left(\nabla\varphi(\boldsymbol{x})-\nabla\varphi(\boldsymbol{x'})\right)^T(\boldsymbol{x}-\boldsymbol{x'})\\
& = & (\boldsymbol{x}-\boldsymbol{x'})^T\nabla^2\varphi(\boldsymbol{\tilde{x}})(\boldsymbol{x}-\boldsymbol{x'}),
\end{eqnarray*}
where $\boldsymbol{\tilde{x}}$ is a point on the line segment between $\boldsymbol{x}$ and $\boldsymbol{x'}$. Therefore, $d_\varphi$ amounts to a Mahalanobis distance parameterized by the Hessian matrix of $\varphi$ which depends on the location of $\boldsymbol{x}$ and $\boldsymbol{x'}$. In this respect, learning $\varphi$ can be seen as learning an infinite number of local Mahalanobis distances. They take a nonparametric approach by assuming $\phi$ to belong to a Reproducing Kernel Hilbert Space $\mathcal{H}_K$ associated to a kernel function $K(\boldsymbol{x},\boldsymbol{x'}) = h(\boldsymbol{x}^T\boldsymbol{x'})$ where $h(z)$ is a strictly convex function (set to $\exp(z)$ in the experiments). This allows the derivation of a representer theorem. Setting $\varphi(\boldsymbol{x}) = \sum_{i=1}^n\alpha_ih(\boldsymbol{x}_i^T\boldsymbol{x})$ leads to the following formulation based on classic positive/negative pairs:
\begin{equation}
\label{eq:bk}
\begin{aligned}
 \min_{\boldsymbol{\alpha}\in\mathbb{R}_+^{n},b} &&& \frac{1}{2}\boldsymbol{\alpha}^T\boldsymbol{K}\boldsymbol{\alpha}\quad+\quad C\sum_{(\boldsymbol{x}_i,\boldsymbol{x}_j)\in \mathcal{S}\cup\mathcal{D}}\ell\left(y_{ij}\left[d_\varphi(\boldsymbol{x}_i,\boldsymbol{x}_j)-b\right]\right),
\end{aligned}
\end{equation}
where $\boldsymbol{K}$ is the Gram matrix, $\ell(t) = \max(0,1-t)$ is the hinge loss and $C$ is the trade-off parameter. Problem \eqref{eq:bk} is solved by a simple subgradient descent approach where each iteration has a linear complexity. Note that \eqref{eq:bk} only has $n+1$ variables instead of $d^2$ in most metric learning formulations, leading to very scalable learning. The downside is that computing the learned distance requires $n$ kernel evaluations, which can be expensive for large datasets. The method is evaluated on clustering problems and exhibits good performance, matching or improving that of other metric learning approaches.

\paragraph{PLML (Wang et al.)}
\label{par:plml}

\citet{Wang2012b} propose PLML,\footnote{Source code available at: \url{http://cui.unige.ch/~wangjun/papers/PLML.zip}} a Parametric Local Metric Learning method where a Mahalanobis metric $d^2_{\boldsymbol{M_i}}$ is learned for each training instance $\boldsymbol{x}_i$:
$$d^2_{\boldsymbol{M}_i}(\boldsymbol{x}_i,\boldsymbol{x}_j) = (\boldsymbol{x}_i-\boldsymbol{x}_j)^T\boldsymbol{M}_i(\boldsymbol{x}_i-\boldsymbol{x}_j).$$
$\boldsymbol{M}_i$ is parameterized to be a weighted linear combination of metric bases $\boldsymbol{M}_{b_1},\dots,\boldsymbol{M}_{b_2}$, where $\boldsymbol{M}_{b_j} \succeq 0$ is associated with an anchor point $\boldsymbol{u}_j$.\footnote{In practice, these anchor points are defined as the means of clusters constructed by the $K$-Means algorithm.} In other words, $\boldsymbol{M}_i$ is defined as:
$$\boldsymbol{M}_i = \sum_{j=1}^m W_{ib_j}\boldsymbol{M}_{b_j},\quad W_{i,b_j} \geq 0,\quad \sum_{j=1}^m W_{ib_j} = 1,$$
where the nonnegativity of the weights ensures that the combination is PSD.
The weight learning procedure is a trade-off between three terms: (i) each point $\boldsymbol{x}$ should be close to its linear approximation $\sum_{j=1}^m W_{ib_j}\boldsymbol{u}_j$, (ii) the weighting scheme should be local (i.e., $W_{ib_j}$ should be large if $\boldsymbol{x}_i$ and $\boldsymbol{u}_i$ are similar), and (iii) the weights should vary smoothly over the data manifold (i.e., similar training instances should be assigned similar weights).\footnote{The weights of a test instance can be learned by optimizing the same trade-off given the weights of the training instances, and simply set to the weights of the nearest training instance.}
Given the weights, the basis metrics $\boldsymbol{M}_{b_1},\dots,\boldsymbol{M}_{b_m}$ are then learned in a large-margin fashion using positive and negative training pairs and Frobenius regularization. In terms of scalability, the weight learning procedure is fairly efficient. However, the metric bases learning procedure requires at each step an eigen-decomposition that scales in $O(d^3)$, making the approach intractable for high-dimensional problems. In practice, PLML performs very well on the evaluated datasets, and is quite robust to overfitting due to its global manifold regularization. However, like LMNN, PLML is sensitive to the relevance of the Euclidean distance to assess the similarity between (anchor) points. Note that PLML has many hyper-parameters but in the experiments the authors use default values for most of them. \citet{Huang2013} propose to regularize the anchor metrics to be low-rank and use alternating optimization to solve the problem.

\paragraph{RFD (Xiong et al.)}
\label{par:rfd}

The originality of the Random Forest Distance \citep{Xiong2012} is to see the metric learning problem as a pair classification problem.\footnote{Source code available at: \url{http://www.cse.buffalo.edu/~cxiong/RFD_Package.zip}} Each pair of examples $(\boldsymbol{x},\boldsymbol{x'})$ is mapped to the following feature space:
$$\phi(\boldsymbol{x},\boldsymbol{x'}) = \left[\begin{array}{c}
|\boldsymbol{x}-\boldsymbol{x'}|\\
\frac{1}{2}(\boldsymbol{x}+\boldsymbol{x'})
\end{array}\right]\in\mathbb{R}^{2d}.$$
The first part of $\phi(\boldsymbol{x},\boldsymbol{x'})$ encodes the relative position of the examples and the second part their absolute position, as opposed to the implicit mapping of the Mahalanobis distance which only encodes relative information. The metric is based on a random forest $F$, i.e.,
$$d_{RFD}(\boldsymbol{x},\boldsymbol{x'}) = F(\phi(\boldsymbol{x},\boldsymbol{x'})) =  \frac{1}{T}\sum_{t=1}^Tf_t(\phi(\boldsymbol{x},\boldsymbol{x'})),$$
where $f_t(\cdot)\in\{0,1\}$ is the output of decision tree $t$. RFD is thus highly nonlinear and is able to implicitly adapt the metric throughout the space: when a decision tree in $F$ selects a node split based on a value of the absolute position part, then the entire sub-tree is specific to that region of $\mathbb{R}^{2d}$. As compared to other local metric learning methods, training is very efficient: each tree takes $O(n\log n)$ time to generate and trees can be built in parallel. A drawback is that the evaluation of the learned metric requires to compute the output of the $T$ trees. 
%The authors derive a generalization bound on the pair classification error --- note that this \red{is} not a bound on the metric itself, nor on the original classification problem.
The experiments highlight the importance of encoding absolute information, and show that RFD outperforms some global and local metric learning methods on several datasets and appears to be quite fast.

\subsection{Metric Learning for Histogram Data}
\label{sec:histogram}

Histograms are feature vectors that lie on the probability simplex $\mathcal{S}^d$. This representation is very common in areas dealing with complex objects, such as natural language processing, computer vision or bioinformatics: each instance is represented as a bag of features, i.e., a vector containing the frequency of each feature in the object. Bags-of(-visual)-words \citep{Salton1975,Li2005} are a common example of such data. We present here three metric learning methods designed specifically for histograms.

\paragraph{$\chi^2$-LMNN (Kedem et al.)}
\label{par:chilmnn}

\citet{Kedem2012} propose $\chi^2$-LMNN, which is based on a simple yet prominent histogram metric, the $\chi^2$ distance \citep{Hafner1995}, defined as
\begin{equation}
\label{eq:chi2}
\chi^2(\boldsymbol{x},\boldsymbol{x'}) = \frac{1}{2}\sum_{i=1}^d\frac{(x^i-x'^i)^2}{x^i+x'^i},
\end{equation}
where $x^i$ denotes the $i^{th}$ feature of $\boldsymbol{x}$.\footnote{The sum in \eqref{eq:chi2} must be restricted to entries that are nonzero in either $\boldsymbol{x}$ or $\boldsymbol{x'}$ to avoid division by zero.} Note that $\chi^2$ is a (nonlinear) proper distance. They propose to generalize this distance with a linear transformation, introducing the following pseudo-distance:
$$\chi^2_{\boldsymbol{L}}(\boldsymbol{x},\boldsymbol{x'}) = \chi^2(\boldsymbol{Lx},\boldsymbol{Lx'}),$$
where $\boldsymbol{L} \in \mathbb{R}^{r\times d}$, with the constraint that $\boldsymbol{L}$ maps any $\boldsymbol{x}$ onto $\mathcal{S}^d$ (the authors show that this can be enforced using a simple trick). The objective function is the same as LMNN\footnote{To be precise, it requires an additional parameter. In standard LMNN, due to the linearity of the Mahalanobis distance, solutions obtained with different values of the margin only differ up to a scaling factor---the margin is thus set to 1. Here, $\chi^2$ is nonlinear and therefore this value must be tuned.} and is optimized using a standard subgradient descent procedure. Although subject to local optima, experiments show great improvements on histogram data compared to standard histogram metrics and Mahalanobis distance learning methods, and promising results for dimensionality reduction (when $r < d$).

\paragraph{GML (Cuturi \& Avis)}
\label{par:gml}

While $\chi^2$-LMNN optimizes a simple bin-to-bin histogram distance, \citet{Cuturi2011} propose to consider the more powerful cross-bin Earth Mover's Distance (EMD) introduced by \citet{Rubner2000}, which can be seen as the distance between a source histogram $\boldsymbol{x}$ and a destination histogram $\boldsymbol{x'}$. On an intuitive level, $\boldsymbol{x}$ is viewed as piles of earth at several locations (bins) and $\boldsymbol{x'}$ as several holes, where the value of each feature represents the amount of earth and the capacity of the hole respectively. The EMD is then equal to the minimum amount of effort needed to move all the earth from $\boldsymbol{x}$ to $\boldsymbol{x'}$. The costs of moving one unit of earth from bin $i$ of $\boldsymbol{x}$ to bin $j$ of $\boldsymbol{x'}$ is encoded in the so-called ground distance matrix $\boldsymbol{D} \in \mathbb{R}^{d\times d}$.\footnote{For EMD to be proper distance, $\boldsymbol{D}$ must satisfy the following $\forall i,j,k\in \{1,\dots,d\}$: (i) $d_{ij} \geq 0$, (ii) $d_{ii} = 0$, (iii) $d_{ij} = d_{ji}$ and (iv) $d_{ij} \leq d_{ik} + d_{kj}$.} The computation of EMD amounts to finding the optimal flow matrix $\boldsymbol{F}$, where $f_{ij}$ corresponds to the amount of earth moved from bin $i$ of $\boldsymbol{x}$ to bin $j$ of $\boldsymbol{x'}$. Given the ground distance matrix $\boldsymbol{D}$, $\text{EMD}_{\boldsymbol{D}}(\boldsymbol{x},\boldsymbol{x'})$ is linear and can be formulated as a linear program:
$$\text{EMD}_{\boldsymbol{D}}(\boldsymbol{x},\boldsymbol{x'}) = \min_{\boldsymbol{f}\in\mathbb{C}(\boldsymbol{x},\boldsymbol{x'})}\boldsymbol{d}^T\boldsymbol{f},$$
where $\boldsymbol{f}$ and $\boldsymbol{d}$ are respectively the flow and the ground matrices rewritten as vectors for notational simplicity, and $\mathbb{C}(\boldsymbol{x},\boldsymbol{x'})$ is the convex set of feasible flows (which can be represented as linear constraints).
Ground Metric Learning (GML) aims at learning $\boldsymbol{D}$ based on training triplets $(\boldsymbol{x}_i,\boldsymbol{x}_j,w_{ij})$ where $\boldsymbol{x}_i$ and $\boldsymbol{x}_j$ are two histograms and $w_{ij}\in\mathbb{R}$ is a weight quantifying the similarity between $\boldsymbol{x}_i$ and $\boldsymbol{x}_j$. The optimized criterion essentially aims at minimizing the sum of $w_{ij}\text{EMD}_{\boldsymbol{D}}(\boldsymbol{x}_i,\boldsymbol{x}_j)$ --- which is a nonlinear function in $\boldsymbol{D}$ --- by casting the problem as a difference of two convex functions. A local minima is found efficiently by a subgradient descent approach. Experiments on image datasets show that GML outperforms standard histogram distances as well as Mahalanobis distance methods.

\paragraph{EMDL (Wang \& Guibas)}
\label{par:emdl}

Building on GML and successful Mahalanobis distance learning approaches such as LMNN, \citet{Wang2012d} aim at learning the EMD ground matrix in the more flexible setting where the algorithm is provided with a set of relative constraints $\mathcal{R}$ that must be satisfied with a large margin. The problem is formulated as
\begin{equation}
\label{eq:edml}
\begin{aligned}
 \min_{\boldsymbol{D}\in\mathbb{D}} &&& \|\boldsymbol{D}\|_{\mathcal{F}}^2\quad+\quad C\sum_{i,j,k}\xi_{ijk}\\
 \text{s.t.} &&& \text{EMD}_{\boldsymbol{D}}(\boldsymbol{x}_i,\boldsymbol{x}_k) - \text{EMD}_{\boldsymbol{D}}(\boldsymbol{x}_i,\boldsymbol{x}_j) \geq 1 - \xi_{ijk} && \forall (\boldsymbol{x}_i,\boldsymbol{x}_j,\boldsymbol{x}_k)\in \mathcal{R},
\end{aligned}
\end{equation}
where $\mathbb{D} = \left\{ \boldsymbol{D}\in\mathbb{R}^{d\times d} : \forall i,j\in \{1,\dots,d\}, d_{ij} \geq 0, d_{ii} = 0\right\}$ and $C\geq 0$ is the trade-off parameter.\footnote{Note that unlike in GML, $\boldsymbol{D}\in\mathbb{D}$ may not be a valid distance matrix. In this case, $\text{EMD}_{\boldsymbol{D}}$ is not a proper distance.} The authors also propose a pair-based formulation. Problem \eqref{eq:edml} is bi-convex and is solved using an alternating procedure: first fix the ground metric and solve for the flow matrices (this amounts to a set of standard EMD problems), then solve for the ground matrix given the flows (this is a quadratic program). The algorithm stops when the changes in the ground matrix are sufficiently small. The procedure is subject to local optima (because \eqref{eq:edml} is not jointly convex) and is not guaranteed to converge: there is a need for a trade-off parameter $\alpha$ between stable but conservative updates (i.e., staying close to the previous ground matrix) and aggressive but less stable updates. Experiments on face verification datasets confirm that EMDL improves upon standard histogram distances and Mahalanobis distance learning methods.

\subsection{Generalization Guarantees for Metric Learning}
\label{sec:genml}

The derivation of guarantees on the generalization performance of the learned model is a wide topic in statistical learning theory \citep{Vapnik1971,Valiant1984}. Assuming that data points are drawn i.i.d. from some (unknown but fixed) distribution $P$, one essentially aims at bounding the deviation of the {\it true risk} of the learned model (its performance on unseen data) from its {\it empirical risk} (its performance on the training sample).\footnote{This deviation is typically a function of the number of training examples and some notion of complexity of the model.}

\begin{figure}[t]
\centering
\includegraphics[width=0.95\textwidth]{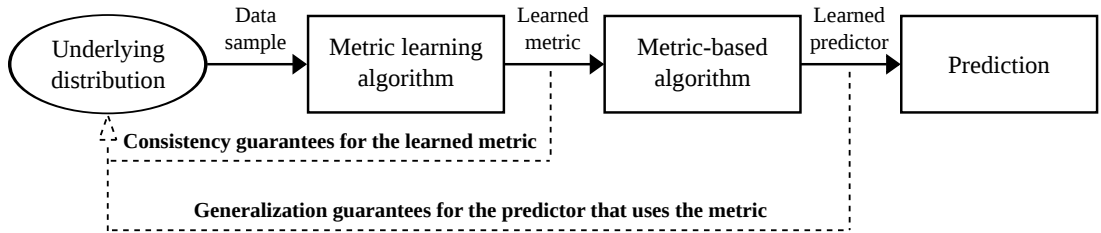}
\caption{The two-fold problem of generalization in metric learning. We may be interested in the generalization ability of the learned metric itself: can we say anything about its consistency on unseen data drawn from the same distribution? Furthermore, we may also be interested in the generalization ability of the predictor using that metric: can we relate its performance on unseen data to the quality of the learned metric?}
\label{fig:generalization}
\end{figure}

In the specific context of metric learning, we claim that the question of generalization can be seen as two-fold \citep{Bellet2012c}, as illustrated by \fref{fig:generalization}:
\begin{itemize}
\item First, one may consider the {\it consistency of the learned metric}, i.e., trying to bound the deviation between the empirical performance of the metric on the training sample and its generalization performance on unseen data.
\item Second, the learned metric is used to improve the performance of some prediction model (e.g., $k$-NN or a linear classifier). It would thus be meaningful to express the {\it generalization performance of this predictor} in terms of that of the learned metric.
\end{itemize}

As in the classic supervised learning setting (where training data consist of individual labeled instances), generalization guarantees may be derived for supervised metric learning (where training data consist of pairs or triplets). Indeed, most of supervised metric learning methods can be seen as minimizing a (regularized) loss function $\ell$ based on the training pairs/triplets.
However, the i.i.d. assumption is violated in the metric learning scenario since the training pairs/triplets are constructed from the training sample. For this reason, establishing generalization guarantees for the learned metric is challenging and only recently has this question been investigated from a theoretical standpoint.

\paragraph{Metric consistency bounds for batch methods} Given a training sample $\mathcal{T} = \{\boldsymbol{z}_i = (\boldsymbol{x}_i,y_i)\}_{i=1}^n$ drawn i.i.d. from an unknown distribution $\mu$, let us consider fully supervised Mahalanobis metric learning of the following general form:
\begin{eqnarray*}
\displaystyle\min_{\boldsymbol{M} \in\mathbb{S}^{d}_+} & \frac{1}{n^2}\displaystyle\sum_{\boldsymbol{z}_i,\boldsymbol{z}_j\in\mathcal{T}}\ell(d_{\boldsymbol{M}}^2,\boldsymbol{z}_i,\boldsymbol{z}_j) \quad+\quad \lambda R(\boldsymbol{M}),
\end{eqnarray*}
where $R(\boldsymbol{M})$ is the regularizer, $\lambda$ the regularization parameter and the loss function $\ell$ is of the form $\ell(d_{\boldsymbol{M}}^2,\boldsymbol{z}_i,\boldsymbol{z}_j) = g(y_iy_j[c-d_{\boldsymbol{M}}^2(\boldsymbol{x}_i,\boldsymbol{x}_j)])$ with $c>0$ a decision threshold variable and $g$ convex and Lipschitz continuous. This includes popular loss functions such as the hinge loss. Several recent work have proposed to study the convergence of the empirical risk (as measured by $\ell$ on pairs from $\mathcal{T}$) to the true risk over the unknown probability distribution $\mu$. The framework proposed by Bian \& Tao \citeyearpar{Bian2011,Bian2012} is quite rigid since it relies on strong assumptions on the distribution of the examples and cannot accommodate any regularization (a constraint to bound $\boldsymbol{M}$ is used instead). \citet{Jin2009} use a notion of uniform stability \citep{Bousquet2002} adapted to the case of metric learning (where training data is made of pairs to derive generalization bounds that are limited to Frobenius norm regularization. \citet{Bellet2012b} demonstrate how to adapt the more flexible notion of algorithmic robustness \citep{Xu2012a} to the metric learning setting to derive (loose) generalization bounds for any matrix norm (including sparsity-inducing ones) as regularizer. They also show that a weak notion of robustness is necessary and sufficient for metric learning algorithms to generalize well. Lastly, \citet{Cao2012a} use a notion of Rademacher complexity \citep{Bartlett2002} dependent on the regularizer to derive bounds for several matrix norms. All these results can easily adapted to non-Mahalanobis linear metric learning formulations.

\paragraph{Regret bound conversion for online methods}

\citet{Wang2012c,Wang2013} deal with the online learning setting. They show that existing proof techniques to convert regret bounds into generalization bounds \citep[see for instance][]{Cesa-Bianchi2008} only hold for univariate loss functions, but derive an alternative framework that can deal with pairwise losses. At each round, the online algorithm receives a new instance and is assumed to pair it with all previously-seen data points. As this is expensive or even infeasible in practice, \citet{Kar2013} propose to use a buffer containing only a bounded number of the most recent instances. They are also able to obtain tighter bounds based on a notion of Rademacher complexity, essentially adapting and extending the work of \citet{Cao2012a}. These results suggest that one can obtain generalization bounds for most/all online metric learning algorithms with bounded regret (such as those presented in \sref{sec:onlineml}).

\paragraph{Link between learned metric and classification performance} The second question of generalization (i.e., at the classifier level) remains an open problem for the most part. To the best of our knowledge, it has only been addressed in the context of metric learning for linear classification. \citet{Bellet2011,Bellet2012,Bellet2012a} rely upon the theory of learning with $(\epsilon,\gamma,\tau)$-good similarity function \citep{Balcan2008a}, which makes the link between properties of a similarity function and the generalization of a linear classifier built from this similarity. Bellet et al. propose to use $(\epsilon,\gamma,\tau)$-goodness as an objective function for metric learning, and show that in this case it is possible to derive generalization guarantees not only for the learned similarity but also for the linear classifier. \citet{Guo2014} extend the results of Bellet et al. to several matrix norms using a Rademacher complexity analysis, based on techniques from \cite{Cao2012a}.

\subsection{Semi-Supervised Metric Learning Methods}
\label{sec:semi}

In this section, we present two categories of metric learning methods that are designed to deal with semi-supervised learning tasks. The first one corresponds to the standard semi-supervised setting, where the learner makes use of unlabeled pairs in addition to positive and negative constraints. The second one concerns approaches which learn metrics to address semi-supervised domain adaptation problems where the learner has access to labeled data drawn according to a source distribution and unlabeled data generated from a different (but related) target distribution.

\subsubsection{Standard Semi-Supervised Setting}

The following metric learning methods leverage the information brought by the set of {\it unlabeled pairs}, i.e., pairs of training examples that do not belong to the sets of positive and negative pairs:
$$\mathcal{U} = \{(\boldsymbol{x}_i,\boldsymbol{x}_j) : i \neq j, (\boldsymbol{x}_i,\boldsymbol{x}_j)\notin \mathcal{S}\cup\mathcal{D}\}.$$

An early approach by \citet{Bilenko2004} combined semi-supervised clustering with metric learning. In the following, we review general metric learning formulations that incorporate information from the set of unlabeled pairs $\mathcal{U}$.

\paragraph{LRML (Hoi et al.)}
\label{par:lrml}

\citet{Hoi2008,Hoi2010} propose to follow the principles of manifold regularization for semi-supervised learning \citep{Belkin2004} by resorting to a weight matrix $\boldsymbol{W}$ that encodes the similarity between pairs of points.\footnote{Source code available at: \url{http://www.ee.columbia.edu/~wliu/}} Hoi et al. construct $\boldsymbol{W}$ using the Euclidean distance as follows:
$$W_{ij} = \left\{ \begin{array}{ll} 1 & \text{if } \boldsymbol{x}_i\in\mathcal{N}(\boldsymbol{x}_j) \text{ or } \boldsymbol{x}_j\in\mathcal{N}(\boldsymbol{x}_i)\\
0 & \text{otherwise}
\end{array}\right.$$
where $\mathcal{N}(\boldsymbol{x}_j)$ denotes the nearest neighbor list of $\boldsymbol{x}_j$. Using $\boldsymbol{W}$, they use the following regularization known as the graph Laplacian regularizer:
$$\frac{1}{2}\sum_{i,j=1}^n d^2_{\boldsymbol{M}}(\boldsymbol{x}_i,\boldsymbol{x}_j)W_{ij} = \trace(\boldsymbol{XLX^TM}),$$
where $\boldsymbol{X}$ is the data matrix and $\boldsymbol{L} = \boldsymbol{D}-\boldsymbol{W}$ is the graph Laplacian matrix with $\boldsymbol{D}$ a diagonal matrix such that $D_{ii} = \sum_j W_{ij}$. Intuitively, this regularization favors an ``affinity-preserving'' metric: the distance between points that are similar according to $\boldsymbol{W}$ should remain small according to the learned metric. Experiments show that LRML (Laplacian Regularized Metric Learning) significantly outperforms supervised methods when the side information is scarce. An obvious drawback is that computing $\boldsymbol{W}$ is intractable for large-scale datasets. This work has inspired a number of extensions and improvements: \citet{Liu2010} introduce a refined way of constructing $\boldsymbol{W}$ while \citet{Baghshah2009}, \citet{Zhong2011} and \citet{Wang2013a} use a different (but similar in spirit) manifold regularizer.

\paragraph{M-DML (Zha et al.)}
\label{par:lmdml}

The idea of \citet{Zha2009} is to augment Laplacian regularization with metrics $\boldsymbol{M}_1,\dots,\boldsymbol{M}_K$ learned from auxiliary datasets. Formally, for each available auxiliary metric, a weight matrix $\boldsymbol{W}_k$ is constructed following \citet{Hoi2008,Hoi2010} but using metric $\boldsymbol{M}_k$ instead of the Euclidean distance. These are then combined to obtain the following regularizer:
$$\sum_{k=1}^K\alpha_k\trace(\boldsymbol{XL}_k\boldsymbol{X^TM}),$$
where $\boldsymbol{L}_k$ is the Laplacian associated with weight matrix $\boldsymbol{W}_k$ and $\alpha_k$ is the weight reflecting the utility of auxiliary metric $\boldsymbol{M}_k$. As such weights are difficult to set in practice, Zha et al. propose to learn them together with the metric $\boldsymbol{M}$ by alternating optimization (which only converges to a local minimum). Experiments on a face recognition task show that metrics learned from auxiliary datasets can be successfully used to improve performance over LRML.

\paragraph{SERAPH (Niu et al.)}
\label{par:seraph}

\citet{Niu2012} tackle semi-supervised metric learning from an information-theoretic perspective by optimizing a probability of labeling a given pair parameterized by a Mahalanobis distance:\footnote{Source code available at: \url{http://sugiyama-www.cs.titech.ac.jp/~gang/software.html}}
$$p^{\boldsymbol{M}}(y | \boldsymbol{x},\boldsymbol{x'}) = \frac{1}{1+\exp\left(y(d^2_{\boldsymbol{M}}(\boldsymbol{x},\boldsymbol{x'}) - \eta)\right)}.$$
$\boldsymbol{M}$ is optimized to maximize the entropy of $p^{\boldsymbol{M}}$ on the labeled pairs $\mathcal{S}\cup\mathcal{D}$ and minimize it on unlabeled pairs $\mathcal{U}$, following the entropy regularization principle \citep{Grandvalet2004}. Intuitively, the regularization enforces low uncertainty of unobserved weak labels. They also encourage a low-rank projection by using the trace norm. The resulting nonconvex optimization problem is solved using an EM-like iterative procedure where the M-step involves a projection on the PSD cone. The proposed method outperforms supervised metric learning methods when the amount of supervision is very small, but was only evaluated against one semi-supervised method \citep{Baghshah2009} known to be subject to overfitting.

\subsubsection{Metric Learning for Domain Adaptation}

In the domain adaptation (DA) setting \citep{Mansour2009,Quinonero-Candela2009,Ben-David2010}, the labeled training data and the test data come from different (but somehow related) distributions (referred to as the source and target distributions respectively). This situation occurs very often in real-world applications---famous examples include speech recognition, spam detection and object recognition---and is also relevant for metric learning. Although domain adaptation is sometimes achieved by using a small sample of labeled target data \citep{Saenko2010,Kulis2011}, we review here the more challenging case where only unlabeled target data is available.

\paragraph{CDML (Cao et al.)}
\label{par:cdml}

CDML \citep{Cao2011}, for Consistent Distance Metric Learning, deals with the setting of covariate shift, which assumes that source and target data distributions $p_S(\boldsymbol{x})$ and $p_T(\boldsymbol{x})$ are different but the conditional distribution of the labels given the features, $p(y | \boldsymbol{x})$, remains the same. In the context of metric learning, the assumption is made at the pair level, i.e., $p(y_{ij} | \boldsymbol{x}_i,\boldsymbol{x}_j)$ is stable across domains. Cao et al. show that if some metric learning algorithm minimizing some training loss $\sum_{(\boldsymbol{x}_i,\boldsymbol{x}_j)\in\mathcal{S}\cup\mathcal{D}}\ell(d_{\boldsymbol{M}}^2,\boldsymbol{x}_i,\boldsymbol{x}_j)$ is asymptotically consistent without covariate shift, then the following algorithm is consistent under covariate shift:
\begin{eqnarray}
\label{eq:cdml}
\displaystyle\min_{\boldsymbol{M} \in\mathbb{S}^{d}_+} & \displaystyle\sum_{(\boldsymbol{x}_i,\boldsymbol{x}_j)\in\mathcal{S}\cup\mathcal{D}}w_{ij}\ell(d_{\boldsymbol{M}}^2,\boldsymbol{x}_i,\boldsymbol{x}_j),& \text{ where } w_{ij} = \frac{p_T(\boldsymbol{x}_i)p_T(\boldsymbol{x}_j)}{p_S(\boldsymbol{x}_i)p_S(\boldsymbol{x}_j)}.
\end{eqnarray}
Problem \eqref{eq:cdml} can be seen as cost-sensitive metric learning, where the cost of each pair is given by the importance weight $w_{ij}$. Therefore, adapting a metric learning algorithm to covariate shift boils down to computing the importance weights, which can be done reliably using unlabeled data \citep{Tsuboi2008}. The authors experiment with ITML and show that their adapted version outperforms the regular one in situations of (real or simulated) covariate shift.

\paragraph{DAML (Geng et al.)}
\label{par:daml}

DAML \citep{Geng2011}, for Domain Adaptation Metric Learning, tackles the general domain adaptation setting. In this case, a classic strategy in DA is to use a term that brings the source and target distribution closer. Following this line of work, Geng et al. regularize the metric using the empirical Maximum Mean Discrepancy \citep[MMD,][]{Gretton2006}, a nonparametric way of measuring the difference in distribution between the source sample $S$ and the target sample $T$:
$$MMD(S,T) = \left\|\frac{1}{|S|}\sum_{i=1}^{|S|}\varphi(\boldsymbol{x}_i)-\frac{1}{|T|}\sum_{i=1}^{|T|}\varphi(\boldsymbol{x'_i})\right\|_\mathcal{H}^2,$$
where $\varphi(\boldsymbol{x})$ is a nonlinear feature mapping function that maps $\boldsymbol{x}$ to the Reproducing Kernel Hilbert Space $\mathcal{H}$. The MMD can be computed efficiently using the kernel trick and can thus be used as a (convex) regularizer in kernelized metric learning algorithms (see \sref{sec:kernelization}). DAML is thus a trade-off between satisfying the constraints on the labeled source data and finding a projection that minimizes the discrepancy between the source and target distribution. Experiments on face recognition and image annotation tasks in the DA setting highlight the effectiveness of DAML compared to classic metric learning methods.

%% file: structured.tex
\section{Metric Learning for Structured Data}
\label{sec:structured}

In many domains, data naturally come structured, as opposed to the ``flat'' feature vector representation we have focused on so far. Indeed, instances can come in the form of strings, such as words, text documents or DNA sequences; trees like XML documents, secondary structure of RNA or parse trees; and graphs, such as networks, 3D objects or molecules. In the context of structured data, metrics are especially appealing because they can be used as a proxy to access data without having to manipulate these complex objects. Indeed, given an appropriate structured metric, one can use any metric-based algorithm as if the data consisted of feature vectors. Many of these metrics actually rely on representing structured objects as feature vectors, such as some string kernels \citep[see][and variants]{Lodhi2002} or bags-of-(visual)-words \citep{Salton1975,Li2005}. In this case, metric learning can simply be performed on the feature vector representation, but this strategy can imply a significant loss of structural information.
On the other hand, there exist metrics that operate directly on the structured objects and can thus capture more structural distortions. However, learning such metrics is challenging because most of structured metrics are combinatorial by nature, which explains why it has received less attention than metric learning from feature vectors. In this section, we focus on the edit distance, which basically measures (in terms of number of operations) the cost of turning an object into another. Edit distance has attracted most of the interest in the context of metric learning for structured data because (i) it is defined for a variety of objects:  sequences \citep{Levenshtein1966}, trees \citep{Bille2005} and graphs \citep{Gao2010}, (ii) it is naturally amenable to learning due to its parameterization by a cost matrix.

We review string edit distance learning in \sref{sec:stringedit}, while methods for trees and graphs are covered in \sref{sec:treeedit}. The features of each approach are summarized in \tref{tab:mlstructsum}.

\begin{table}[t]
\centering
\begin{scriptsize}
\begin{tabular}{ccccccccc}
\toprule
\multirow{2}{*}{\textbf{Page}}&\multirow{2}{*}{\textbf{Name}} & \multirow{2}{*}{\textbf{Year}} & \textbf{Source} & \textbf{Data} & \multirow{2}{*}{\textbf{Method}} & \multirow{2}{*}{\textbf{Script}} & \multirow{2}{*}{\textbf{Optimum}} & \textbf{Negative}\\
& & & \textbf{Code} & \textbf{Type} &&&& \textbf{Pairs}\\
\midrule
\pageref{par:Ristad}&R\&Y & 1998 & Yes & String & Generative+EM & All & Local & No\\
\pageref{par:Oncina}&O\&S & 2006 & Yes & String & Discriminative+EM & All & Local & No\\
\pageref{par:Saigo}&Saigo & 2006 & Yes & String & Gradient Descent & All & Local & No\\
\pageref{par:GESL}&GESL & 2011 & Yes & All & Gradient Descent &Levenshtein&Global&Yes\\
\pageref{par:Bernard}&Bernard & 2006 & Yes & Tree & Both+EM & All  & Local & No\\
\pageref{par:Boyer}&Boyer & 2007 & Yes & Tree & Generative+EM & All  & Local & No\\
\pageref{par:Dalvi}&Dalvi & 2009 & No & Tree & Discriminative+EM & All  & Local & No\\
\pageref{par:Emms}&Emms & 2012 & No & Tree & Discriminative+EM & Optimal  & Local & No\\
\pageref{par:Neuhaus}&N\&B & 2007 & No & Graph & Generative+EM & All  & Local & No\\
\bottomrule
\end{tabular}
\end{scriptsize}
\caption{Main features of metric learning methods for structured data. Note that all methods make use of positive pairs.}
\label{tab:mlstructsum}
\end{table}

\subsection{String Edit Distance Learning}
\label{sec:stringedit}

In this section, we first introduce some notations as well as the string edit distance. We then review the relevant metric learning methods.

\subsubsection{Notations and Definitions}

\begin{definition}[Alphabet and string]
An alphabet $\Sigma$ is a finite nonempty set of symbols. A string $\mathsf{x}$ is a finite sequence of symbols from $\Sigma$. The empty string/symbol is denoted by $\$$ and $\Sigma^*$ is the set of all finite strings (including $\$$) that can be generated from $\Sigma$. Finally, the length of a string $\mathsf{x}$ is denoted by $|\mathsf{x}|$.
\end{definition}

\begin{definition}[String edit distance]
Let  $\boldsymbol{C}$ be a nonnegative $(|\Sigma|+1)\times (|\Sigma|+1)$ matrix giving the cost of the following elementary edit operations: insertion, deletion and substitution of a symbol, where symbols are taken from $\Sigma\cup \{\$\}$. Given two strings $\mathsf{x},\mathsf{x'}\in\Sigma^*$, an edit script is a sequence of operations that turns $\mathsf{x}$ into $\mathsf{x'}$. The string edit distance \citep{Levenshtein1966} between $\mathsf{x}$ and $\mathsf{x'}$ is defined as the cost of the cheapest edit script and can be computed in $O(|\mathsf{x}|\cdot|\mathsf{x'}|)$ time by dynamic programming.
\end{definition}

Similar metrics include the Needleman-Wunsch score \citep{Needleman1970} and the Smith-Waterman score \citep{Smith1981}. These alignment-based measures use the same substitution operations as the edit distance, but a linear gap penalty function instead of insertion/deletion costs.

The standard edit distance, often called Levenshtein edit distance, is based on a unit cost for all operations. However, this might not reflect the reality of the considered task: for example, in typographical error correction, the probability that a user hits the Q key instead of W on a QWERTY keyboard is much higher than the probability that he hits Q instead of Y. For some applications, such as protein alignment or handwritten digit recognition, hand-tuned cost matrices may be available \citep{Dayhoff1978,Henikoff1992,Mico1998}. Otherwise, there is a need for automatically learning the cost matrix $\boldsymbol{C}$ for the task at hand.

\subsubsection{Stochastic String Edit Distance Learning}
\label{sec:stochedit}

\begin{figure}[t]
  \centering
  \includegraphics[width=0.4\textwidth]{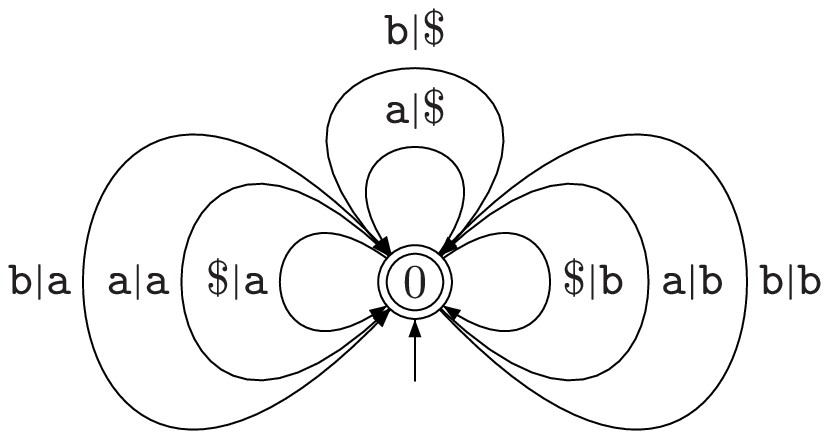}
  \caption[An example of memoryless cPFT]{A memoryless stochastic transducer that models the edit probability of any pair of strings built from $\Sigma=\{\mathtt{a},\mathtt{b}\}$. Edit probabilities assigned to each transition are not shown here for the sake of readability.}
  \label{fig:memoryless}
\end{figure}

Optimizing the edit distance is challenging because the optimal sequence of operations depends on the edit costs themselves, and therefore updating the costs may change the optimal edit script. Most general-purpose approaches get round this problem by considering a stochastic variant of the edit distance, where the cost matrix defines a probability distribution over the edit operations. One can then define an edit similarity as the posterior probability $p_e(\mathsf{x'}|\mathsf{x})$ that an input string $\mathsf{x}$ is turned into an output string $\mathsf{x'}$. This corresponds to summing over all possible edit scripts that turn $\mathsf{x}$ into $\mathsf{x'}$ instead of only considering the optimal script. Such a stochastic edit process can be represented as a probabilistic model, such as a stochastic transducer (\fref{fig:memoryless}), and one can estimate the parameters of the model (i.e., the cost matrix) that maximize the expected log-likelihood of positive pairs. This is done via an EM-like iterative procedure \citep{Dempster1977}. Note that unlike the standard edit distance, the obtained edit similarity does not usually satisfy the properties of a distance (in fact, it is often not symmetric and rarely satisfies the triangular inequality).

\paragraph{Ristad and Yianilos}
\label{par:Ristad}
The first method for learning a string edit metric, in the form of a generative model, was proposed by \citet{Ristad1998}.\footnote{\label{foot:sedil}An implementation is available within the SEDiL platform \citep{Boyer2008}:\\ \url{http://labh-curien.univ-st-etienne.fr/SEDiL/}} They use a memoryless stochastic transducer which models the joint probability of a pair $p_e(\mathsf{x},\mathsf{x'})$ from which $p_e(\mathsf{x'}|\mathsf{x})$ can be estimated. Parameter estimation is performed with an EM procedure. The Expectation step takes the form of a probabilistic version of the dynamic programing algorithm of the standard edit distance. The M-step aims at maximizing the likelihood of the training pairs of strings so as to define a joint distribution over the edit operations:%, under the following constraint:
\begin{eqnarray*}
\sum_{(u,v) \in (\Sigma \cup \{\$\})^2\backslash\{\$,\$\}}\boldsymbol{C}_{uv}+c(\#)=1,\quad\text{ with } c(\#) > 0 \text{ and } \boldsymbol{C}_{uv} \geq 0,
\end{eqnarray*}
where $\#$ is a termination symbol and $c(\#)$ the associated cost (probability). %The learned edit probability has been efficiently applied to the problem of learning word pronunciation in conversational speech.\\

Note that \citet{Bilenko2003} extended this approach to the Needleman-Wunsch score with affine gap penalty and applied it to duplicate detection. To deal with the tendency of Maximum Likelihood estimators to overfit when the number of parameters is large (in this case, when the alphabet size is large), \citet{Takasu2009} proposes a Bayesian parameter estimation of pair-HMM providing a way to smooth the estimation. %Experiments are conducted on approximate text searching in a digital library of Japanese and English documents.

\paragraph{Oncina and Sebban}
\label{par:Oncina}
The work of \citet{Oncina2006} describes three levels of bias induced by the use of generative models: (i) dependence between edit operations, (ii) dependence between the costs and the prior distribution of strings $p_e(\mathsf{x})$, and (iii) the fact that to obtain the posterior probability one must divide by the empirical estimate of $p_e(\mathsf{x})$. These biases are highlighted by empirical experiments conducted with the method of \citet{Ristad1998}. To address these limitations, they propose the use of a conditional transducer as a discriminative model that directly models the posterior probability $p(\mathsf{x'}|\mathsf{x})$ that an input string $\mathsf{x}$ is turned into an output string $\mathsf{x'}$ using edit operations.\cref{foot:sedil} Parameter estimation is also done with EM where the maximization step differs from that of \citet{Ristad1998} as shown below:
\begin{eqnarray*}
\forall u \in \Sigma, \sum_{v \in \Sigma \cup \{\$\}}\boldsymbol{C}_{v|u}+ \sum_{v \in \Sigma}\boldsymbol{C}_{v|\$}=1,\quad\text{ with } \sum_{v \in \Sigma}\boldsymbol{C}_{v|\$}+c(\#)=1.
\end{eqnarray*}

%The paper features an application to handwritten digit recognition, where digits are represented as sequences of Freeman codes \citep{Freeman1974}.\\
In order to allow the use of negative pairs, \citet{McCallum2005} consider another discriminative model, conditional random fields, that can deal with positive and negative pairs in specific states, still using EM for parameter estimation.

\subsubsection{String Edit Distance Learning by Gradient Descent}

The use of EM has two main drawbacks: (i) it may converge to a local optimum, and (ii) parameter estimation and distance calculations must be done at each iteration, which can be very costly if the size of the alphabet and/or the length of the strings are large. 
The following methods get round these drawbacks by formulating the learning problem in the form of an optimization problem that can be efficiently solved by a gradient descent procedure.

\paragraph{Saigo et al.}
\label{par:Saigo} 

\citet{Saigo2006} manage to avoid the need for an iterative procedure like EM in the context of detecting remote homology in protein sequences.\footnote{Source code available at: \url{http://sunflower.kuicr.kyoto-u.ac.jp/~hiroto/project/optaa.html}} They learn the parameters of the Smith-Waterman score which is plugged in their local alignment kernel $k_{LA}$ where all the possible local alignments $\pi$ for changing $\mathsf{x}$ into $\mathsf{x}'$ are taken into account~\citep{Saigo2004}:
\begin{equation}
\label{LA-kernel}
k_{LA}(x,x')=\sum_\pi e^{t \cdot s(x,x',\pi)}.
\end{equation}
In the above formula, $t$ is a parameter and $s(x,x',\pi)$ is the corresponding score of $\pi$ and defined as follows:
\begin{equation}
\label{LA-similarity}
s(x,x',\pi)=\sum_{u,v\in\Sigma}n_{u,v}(x,x',\pi) \cdot \boldsymbol{C}_{uv}-n_{g_d}(x,x',\pi) \cdot g_d-n_{g_e}(x,x',\pi) \cdot g_e,
\end{equation}
where $n_{u,v}(x,x',\pi)$ is the number of times that symbol $u$ is aligned with $v$ while $g_d$ and $g_e$, along with their corresponding number of occurrences $n_{g_d}(x,x',\pi)$ and $n_{g_e}(x,x',\pi)$, are two parameters dealing respectively with the opening and extension of gaps.

Unlike the Smith-Waterman score, $k_{LA}$ is differentiable and can be optimized by a gradient descent procedure. The objective function that they optimize is meant to favor the discrimination between positive and negative examples, but this is done by only using positive pairs of distant homologs. The approach has two additional drawbacks: (i) the objective function is nonconvex and thus subject to local minima, and (ii) in general, $k_{LA}$ does not fulfill the properties of a kernel.

\paragraph{GESL (Bellet et al.)}
\label{par:GESL} 
\citet{Bellet2011,Bellet2012} propose a convex programming approach to learn edit similarity functions from both positive and negative pairs without requiring a costly iterative procedure.\footnote{Source code available at: \url{http://www-bcf.usc.edu/~bellet/}} They use the following simplified edit function: 
$$e_{\boldsymbol{C}}(\mathsf{x},\mathsf{x'}) = \displaystyle\sum_{(u,v) \in (\Sigma \cup \{\$\})^2\backslash\{\$,\$\}}\boldsymbol{C}_{uv}\cdot \#_{uv}(\mathsf{x},\mathsf{x'}),$$
where $\#_{uv}(\mathsf{x},\mathsf{x'})$ is the number of times the operation $u\to v$ appears in the Levenshtein script. Therefore, $e_{\boldsymbol{C}}$ can be optimized directly since the sequence of operations is fixed (it does not depend on the costs). The authors optimize the nonlinear similarity $K_{\boldsymbol{C}}(\mathsf{x},\mathsf{x'}) = 2\exp(-e_{\boldsymbol{C}}(\mathsf{x},\mathsf{x'}))-1$, derived from $e_{\boldsymbol{C}}$. Note that $K_{\boldsymbol{C}}$ is not required to be PSD nor symmetric. 
GESL (Good Edit Similarity Learning) is expressed as follows:
\begin{equation}
\label{eq:gesl}
\begin{aligned}
\min_{\boldsymbol{C},B_1,B_2} &&&\frac{1}{n^2}\displaystyle\sum_{z_i,z_j}\ell(\boldsymbol{C},z_i,z_j)\quad+\quad\beta\|\boldsymbol{C}\|^2_{\cal{F}}\\
\text{s.t.} &&& B_1 \geq -\log(\frac{1}{2}),\quad 0 \leq B_2 \leq -\log(\frac{1}{2}),\quad B_1 - B_2 = \eta_\gamma\nonumber,
\end{aligned}
\end{equation}
where $\beta \geq 0$ is a regularization parameter, $\eta_\gamma \geq 0$ a parameter corresponding to a desired ``margin'' and
$$\ell(\boldsymbol{C},z_i,z_j)=
\left\{\begin{array}{l}
{[} B1 - e_{\boldsymbol{C}}(\mathsf{x_i},\mathsf{x_j}) {]}_{+} \textrm{ if } y_i\neq y_j\\
{[} e_{\boldsymbol{C}}(\mathsf{x_i},\mathsf{x_j}) - B2 {]}_{+} \textrm{ if } y_i=y_j .
\end{array}\right.$$
GESL essentially learns the edit cost matrix $\boldsymbol{C}$ so as to optimize the $(\epsilon,\gamma,\tau)$-goodness \citep{Balcan2008a} of the similarity $K_{\boldsymbol{C}}(\mathsf{x},\mathsf{x'})$ and thereby enjoys generalization guarantees both for the learned similarity and for the resulting linear classifier (see \sref{sec:genml}). A potential drawback of GESL is that it optimized a simplified variant of the edit distance, although this does not seem to be an issue in practice. Note that GESL can be straightforwardly adapted to learn tree or graph edit similarities \citep{Bellet2012}.

\subsection{Tree and Graph Edit Distance Learning}
\label{sec:treeedit}

In this section, we briefly review the main approaches in tree/graph edit distance learning. We do not delve into the details of these approaches as they are essentially adaptations of stochastic string edit distance learning presented in \sref{sec:stochedit}.

\paragraph{Bernard et al.}\label{par:Bernard} Extending the work of \citet{Ristad1998} and \citet{Oncina2006} on string edit similarity learning, \citet{Bernard2006,Bernard2008} propose both a generative and a discriminative model for learning tree edit costs.\cref{foot:sedil} They rely on the tree edit distance by \citet{Selkow1977}---which is cheaper to compute than that of \citet{Zhang1989}---and adapt the updates of EM to this case.

\paragraph{Boyer et al.}\label{par:Boyer} The work of \citet{Boyer2007} tackles the more complex variant of the tree edit distance \citep{Zhang1989}, which allows the insertion and deletion of single nodes instead of entire subtrees only.\cref{foot:sedil} Parameter estimation in the generative model is also based on EM.

\paragraph{Dalvi et al.}\label{par:Dalvi} The work of \citet{Dalvi2009} points out a limitation of the approach of \citet{Bernard2006,Bernard2008}: they model a distribution over tree edit scripts rather than over the trees themselves, and unlike the case of strings, there is no bijection between the edit scripts and the trees. Recovering the correct conditional probability with respect to trees requires a careful and costly procedure. They propose a more complex conditional transducer that models the conditional probability over trees and use again EM for parameter estimation.

\paragraph{Emms}\label{par:Emms} The work of \citet{Emms2012} points out a theoretical limitation of the approach of \citet{Boyer2007}: the authors use a factorization that turns out to be incorrect in some cases. Emms shows that a correct factorization exists when only considering the edit script of highest probability instead of all possible scripts, and derives the corresponding EM updates. An obvious drawback is that the output of the model is not the probability $p(\mathsf{x'}|\mathsf{x})$. Moreover, the approach is prone to overfitting and requires smoothing and other heuristics (such as a final step of zeroing-out the diagonal of the cost matrix).

\paragraph{Neuhaus \& Bunke}\label{par:Neuhaus} In their paper, \citet{Neuhaus2007} learn a (more general) graph edit similarity, where each edit operation is modeled by a Gaussian mixture density. Parameter estimation is done using an EM-like algorithm. Unfortunately, the approach is intractable: the complexity of the EM procedure is exponential in the number of nodes (and so is the computation of the distance).

%% file: conclu.tex
\section{Conclusion and Discussion}
\label{sec:conclu}

In this survey, we provided a comprehensive review of the main methods and trends in metric learning. We here briefly summarize and draw promising lines for future research.

\subsection{Summary}

\paragraph{Numerical data} While metric learning for feature vectors was still in its early life at the time of the first survey \citep{Yang2006}, it has now reached a good maturity level. Indeed, recent methods are able to deal with a large spectrum of settings in a scalable way. In particular, online approaches have played a significant role towards better scalability, complex tasks can be tackled through nonlinear or local metric learning, methods have been derived for difficult settings such as ranking, multi-task learning or domain adaptation, and the question of generalization in metric learning has been the focus of recent papers.

\paragraph{Structured data}
On the other hand, much less work has gone into metric learning for structured data and advances made for numerical data have not yet propagated to structured data. Indeed, most approaches remain based on EM-like algorithms which make them intractable for large datasets and instance size, and hard to analyze due to local optima. Nevertheless, recent advances such as GESL \citep{Bellet2011} have shown that drawing inspiration from successful feature vector formulations (even if it requires simplifying the metric) can be highly beneficial in terms of scalability and flexibility. This is promising direction and probably a good omen for the development of this research area.

\subsection{What next?}

In light of this survey, we can identify the limitations of the current literature and speculate on where the future of metric learning is going.

\paragraph{Scalability with both $n$ and $d$} There has been satisfying solutions to perform metric learning on large datasets (``Big Data'') through online learning or stochastic optimization. The question of scalability with the dimensionality is more involved, since most methods learn $O(d^2)$ parameters, which is intractable for real-world applications involving thousands of features, unless dimensionality reduction is applied beforehand. Kernelized methods have $O(n^2)$ parameters instead, but this is infeasible when $n$ is also large. Therefore, the challenge of achieving high scalability with both $n$ and $d$ has yet to be overcome.
Recent approaches have tackled the problem by optimizing over the manifold of low-rank matrices \citep{Shalit2012,Cheng2013} or defining the metric based on a combination of simple classifiers \citep{Kedem2012,Xiong2012}. These approaches have a good potential for future research.

\paragraph{More theoretical understanding}
Although several recent papers have looked at the generalization of metric learning, analyzing the link between the consistency of the learned metric and its performance in a given algorithm (classifier, clustering procedure, etc) remains an important open problem. So far, only results for linear classification have been obtained \citep{Bellet2012a,Guo2014}, while learned metrics are also heavily used for $k$-NN classification, clustering or information retrieval, for which no theoretical result is known.

\paragraph{Unsupervised metric learning} A natural question to ask is whether one can learn a metric in a purely unsupervised way. So far, this has only been done as a byproduct of dimensionality reduction algorithms. Other relevant criteria should be investigated, for instance learning a metric that is robust to noise or invariant to some transformations of interest, in the spirit of denoising autoencoders \citep{Vincent2008,Chen2012}. Some results in this direction have been obtained for image transformations \citep{Kumar2007}. A related problem is to characterize what it means for a metric to be good for clustering. There has been preliminary work on this question \citep{Balcan2008b,Lajugie2014}, which deserves more attention.

\paragraph{Leveraging the structure} The simple example of metric learning designed specifically for histogram data \citep{Kedem2012} has shown that taking the structure of the data into account when learning the metric can lead to significant improvements in performance. As data is becoming more and more structured (e.g., social networks), using this structure to bias the choice of metric is likely to receive increasing interest in the near future.

\paragraph{Adapting the metric to changing data}
An important issue is to develop methods robust to changes in the data. In this line of work, metric learning in the presence of noisy data as well as for transfer learning and domain adaptation have recently received some interest. However, these efforts are still insufficient for dealing with lifelong learning applications, where the learner experiences concept drift and must detect and adapt the metric to different changes.

\paragraph{Learning richer metrics}
Existing metric learning algorithms ignore the fact that the notion of similarity is often multimodal: there exist several ways in which two instances may be similar (perhaps based on different features), and different degrees of similarity (versus the simple binary similar/dissimilar view). Being able to model these shades as well as to interpret why things are similar would bring the learned metrics closer to our own notions of similarity.